\documentclass[10pt,twocolumn,letterpaper]{article}

\usepackage{iccv}
\usepackage{times}
\usepackage{epsfig}
\usepackage{graphicx}
\usepackage{subcaption}
\usepackage{amsmath}
\usepackage{amssymb}
\usepackage{amsmath,amsthm,amssymb,bm}
\usepackage{amsfonts}
\usepackage{mathrsfs}
\usepackage{multirow}
\usepackage{xspace}
\usepackage{array}
\usepackage{enumerate}
\usepackage{algorithm}
\usepackage{algpseudocode}
\usepackage{appendix}
\usepackage{wrapfig}
\usepackage{sidecap}
\usepackage{booktabs}
\usepackage{booktabs}       
\usepackage{amsfonts}       
\usepackage{nicefrac}       
\usepackage{microtype}      
\usepackage{ifthen}
\usepackage[font=small,labelfont=bf]{caption}
\usepackage{booktabs}       
\usepackage{amsfonts}       
\usepackage{nicefrac}       
\usepackage{microtype}      
\usepackage[export]{adjustbox} 
\usepackage{tabularx, booktabs, ragged2e}

\usepackage[pagebackref=true,breaklinks=true,letterpaper=true,colorlinks,bookmarks=false]{hyperref}

\usepackage[capitalize]{cleveref}
\usepackage{authblk}

\DeclareMathOperator{\softmax}{softmax}

\usepackage{soul}

\usepackage{amsthm}
\newtheorem{remark}{Remark}

\usepackage{color}
\definecolor{darkgreen}{rgb}{0,0.5,0}
\definecolor{purple}{rgb}{1,0,1}
\newcommand{\comm}[2]{\ifnum\COMMENTs=1\textcolor{#1}{#2}\fi}

\iccvfinalcopy 

\def\iccvPaperID{3774} 

\ificcvfinal\fi
\begin{document}

\setlength{\abovedisplayskip}{3pt}
\setlength{\belowdisplayskip}{3pt}

\title{Task-Discriminative Domain Alignment for Unsupervised Domain Adaptation\vspace{-1.5em}}
\author[1]{Behnam Gholami 
}
\author[1]{Pritish Sahu
}
\author[2]{Minyoung Kim
}
\author[1,2]{Vladimir Pavlovic
}
\affil[1]{Dept. of Computer Science, Rutgers University, NJ, USA}
\affil[2]{Samsung AI Center, Cambridge, UK}
\affil[ ]{\normalsize{ \texttt{\{bb510,ps851,vladimir\}@cs.rutgers.edu}, \texttt{v.pavlovic@samsung.com}, \texttt{mikim21@gmail.com}}
\vspace{-1.5em}
}


\maketitle

\begin{abstract}

Domain Adaptation (\textbf{DA}), the process of effectively adapting task models learned on one domain, the source, to other related but distinct domains, the targets, with no or minimal retraining, is typically accomplished using the process of source-to-target manifold alignment. However, this process often leads to unsatisfactory adaptation performance, in part because it ignores the task-specific structure of the data. In this paper, we improve the performance of \textbf{DA} by introducing a discriminative discrepancy measure which takes advantage of auxiliary information available in the source and the target domains to better align the source and target distributions. Specifically, we leverage the cohesive clustering structure within individual data manifolds, associated with different tasks, to improve the alignment. This structure is explicit in the source, where the task labels are available, but is implicit in the target, making the problem challenging. We address the challenge by devising a deep \textbf{DA} framework, which combines a new task-driven domain alignment discriminator with domain regularizers that encourage the shared features as task-specific and domain invariant, and prompt the task model to be data structure preserving, guiding its decision boundaries through the low density data regions. We validate our framework on standard benchmarks, including Digits (\textbf{MNIST}, \textbf{USPS}, \textbf{SVHN}, \textbf{MNIST-M}), \textbf{PACS}, and \textbf{VisDA}. Our results show that our proposed model consistently outperforms the state-of-the-art in unsupervised domain adaptation.

\end{abstract}

\section{Introduction}

Domain adaptation refers to the problem of leveraging labeled task data in a source domain to learn an accurate model of the same tasks in a target domain where the labels are unavailable or very scarce~\cite{csurka2017comprehensive}. The problem becomes challenging in the presence of strong data distribution shifts across the two domains~\cite{sener2016learning,french2018selfensembling}, which lead to high generalization error when using models trained on the source for predicting on target samples. Domain adaptation techniques seek to address the distribution shift problem. The key idea is to bridge the gap between the source and target in a joint feature space so that a task classifier trained on labeled source data can be effectively transferred to the target~\cite{morerio2018minimalentropy,bousmalis2016domain,blitzer2011domain,ming2015unsupervised}. In this regard, an important challenge is how to measure the discrepancy between the two domains. Many domain discrepancy measures have been proposed in previous \textbf{DA} studies, such as the moment matching-based methods~\cite{long2016unsupervised, bousmalis2016domain,pan2011domain,zellinger2017central,yan2017mind}, and adversarial methods~\cite{tzeng2017adversarial,bousmalis2017unsupervised,sankaranarayanan2018generate,zhang2018collaborative,ganin2016domain}. Moment matching-based methods use Maximum Mean Discrepancy (\textbf{MMD})~\cite{sriperumbudur2010hilbert} to align the distributions by matching all their statistics. Inspired by Generative Adversarial Networks (GAN)~\cite{goodfellow2014generative}, adversarial divergences train a domain discriminator to discern the source from the target, while an encoder feature extractor is simultaneously learned to create features indistinguishable across the source and the target, confusing the discriminator.

\begin{figure*}[htb]
\centering\setlength\tabcolsep{1.5pt}
     \begin{tabular}{cc} 
\vspace{-0.5em}
    \begin{subfigure}[b]{0.45\linewidth}
    \includegraphics[trim={0.2cm 2.0cm 0cm 0.7cm},clip,width = \linewidth]{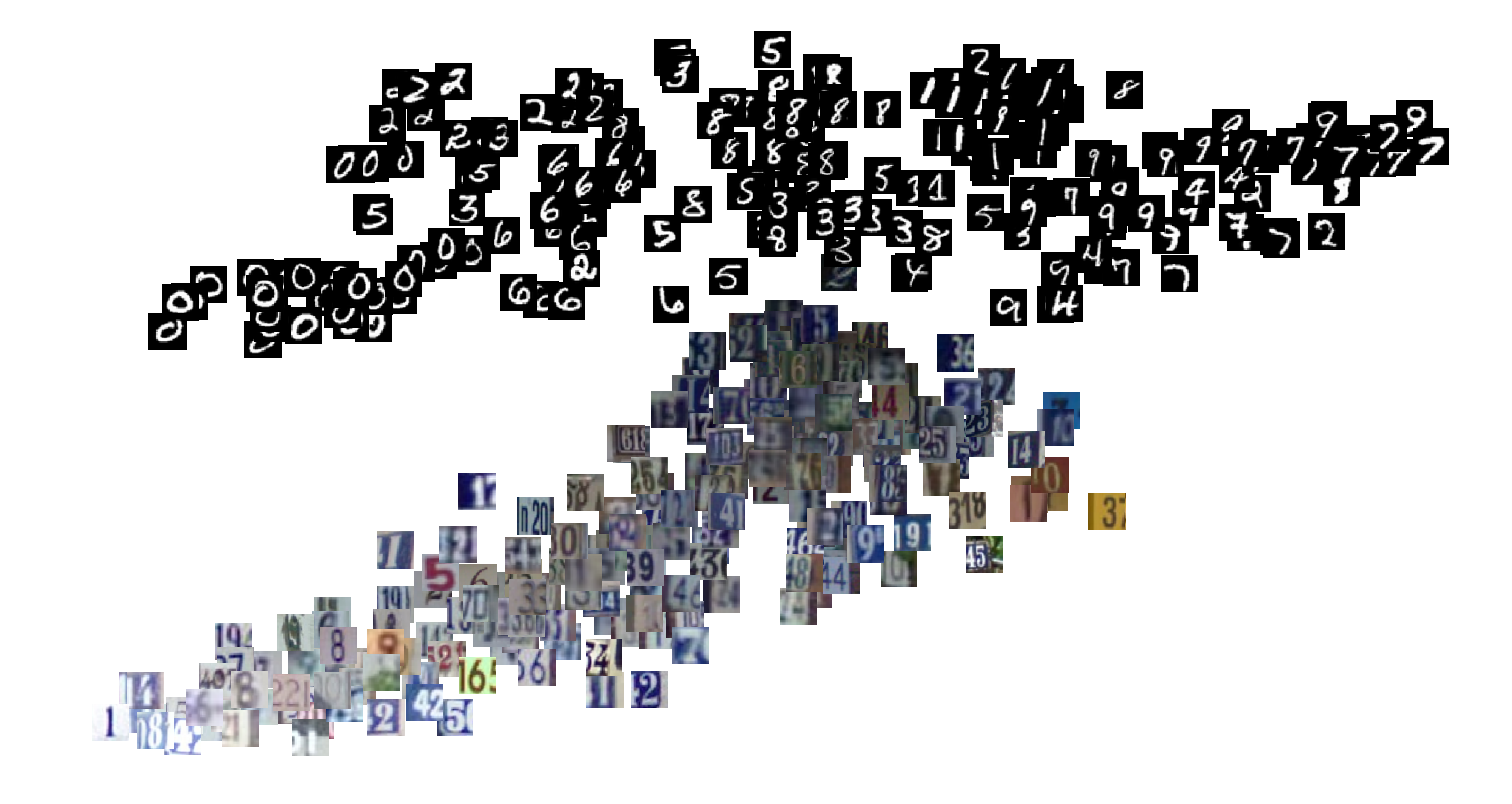}
    \caption{\textbf{Original}\label{r1}}
    \end{subfigure}
    &
    \begin{subfigure}[b]{0.45\linewidth}
	 \includegraphics[trim={0.2cm 2.0cm 0cm 0.7cm},clip,width = \linewidth]{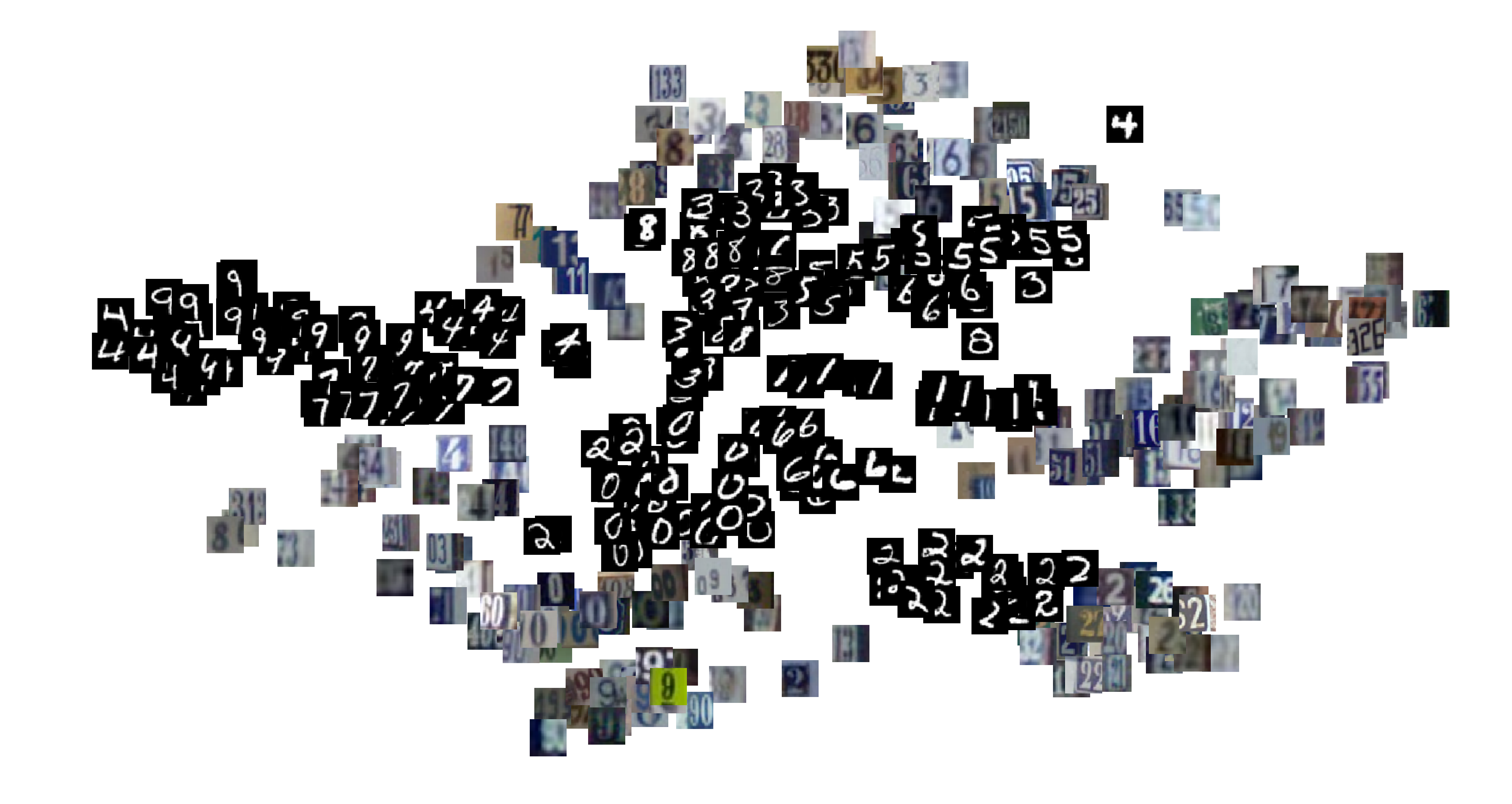}
	 \caption{\textbf{MMD distance}~\cite{long2015learning}
	 \label{r2}}
	 \end{subfigure}
	\\
	 \vspace{-0.8em}
	 \begin{subfigure}[b]{0.45\linewidth}
       \includegraphics[trim={0.2cm 2.0cm 0cm 0.7cm},clip,width = \linewidth]{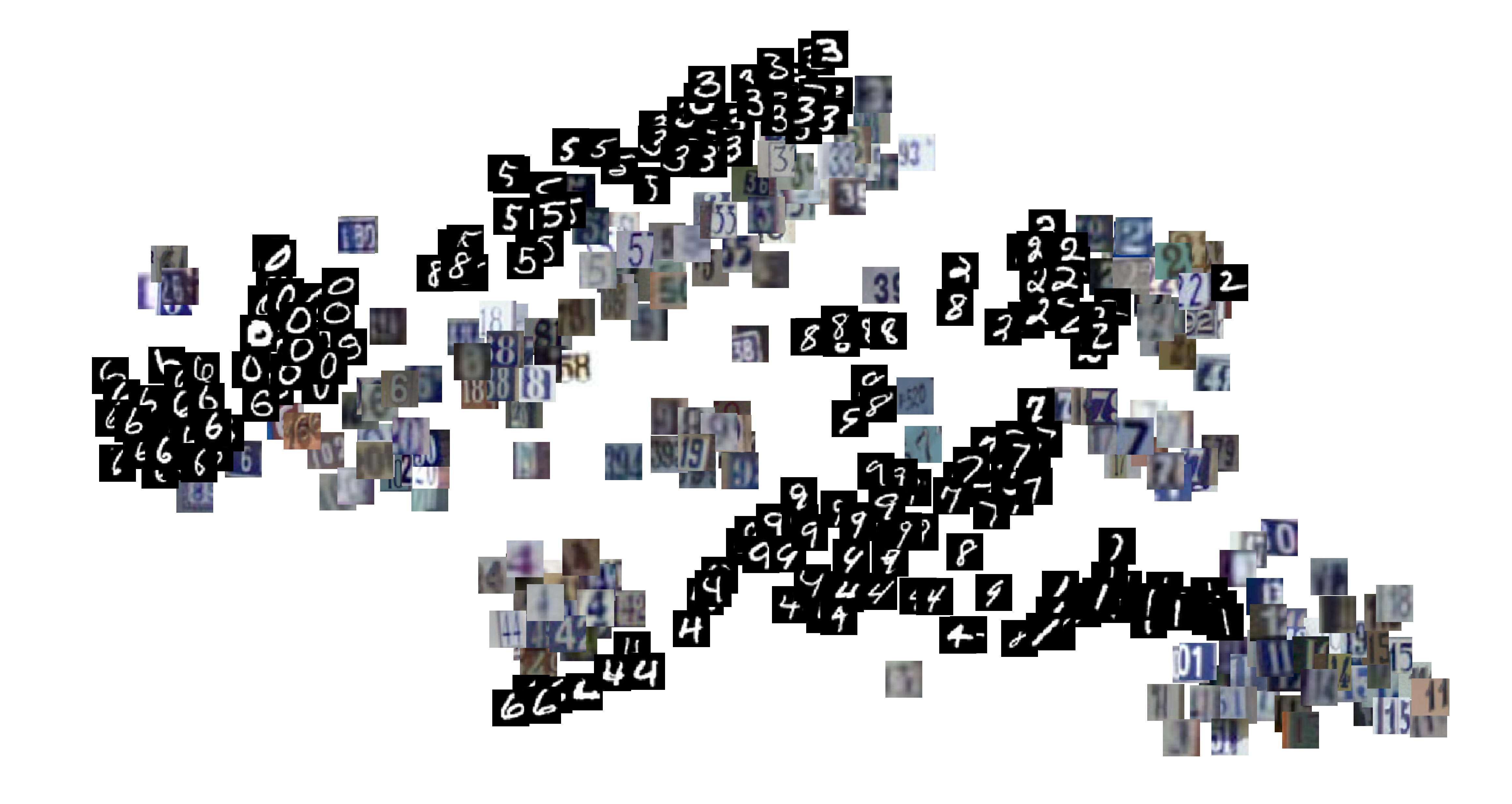}
       \caption{\textbf{Adversarial Divergence} ~\cite{ganin2016domain}\label{r3}}
     \end{subfigure}
     & 
	\begin{subfigure}[b]{0.45\linewidth}
	\includegraphics[trim={0.2cm 2.0cm -0cm 0.7cm},clip,width = \linewidth
	]{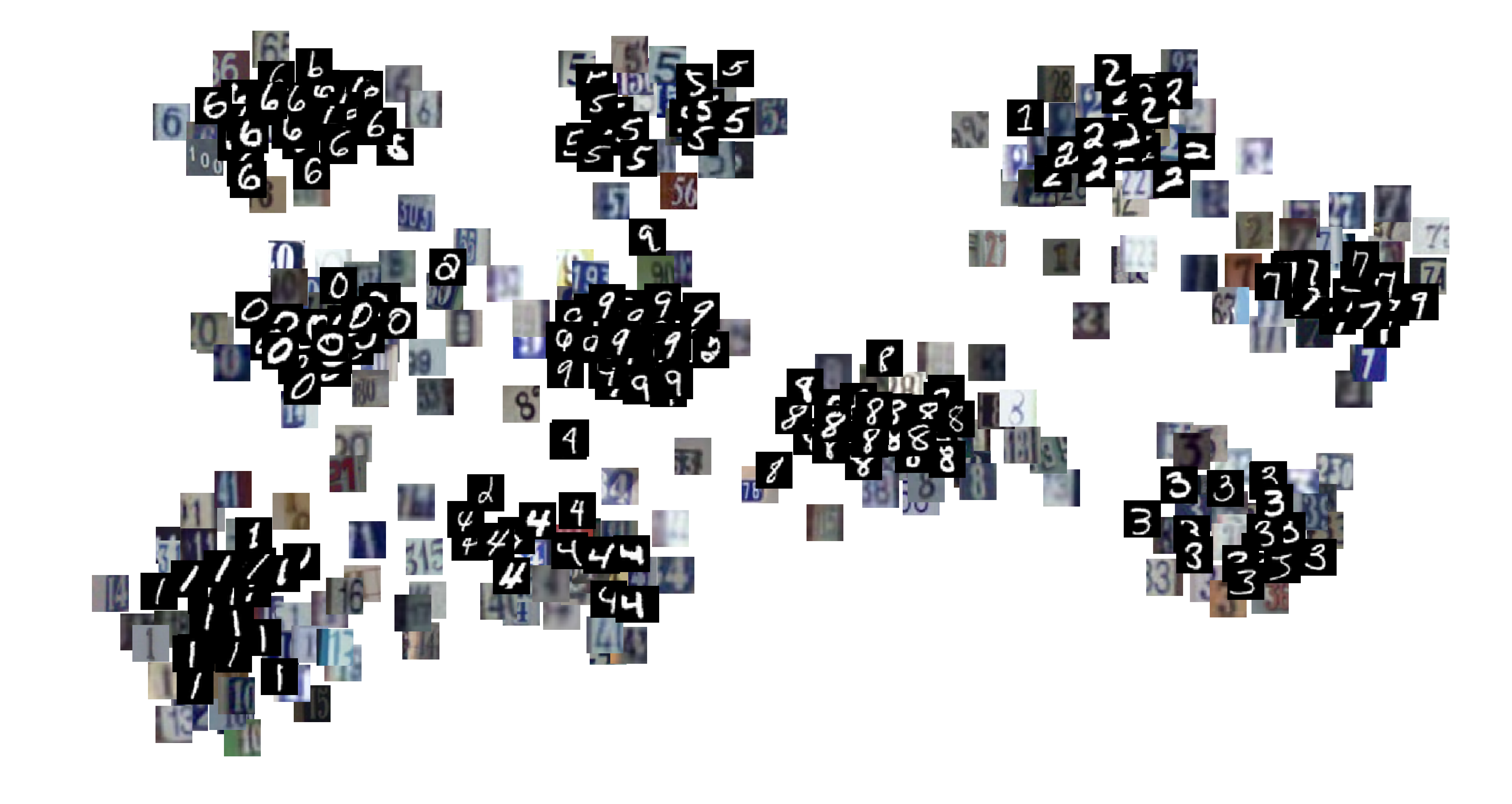}
	\caption{\textbf{Task-specific Divergence}\label{r4}}
	\end{subfigure}
    \end{tabular}
    \caption{Feature visualization, via \textbf{t-SNE}, of Digit datasets when adapting \textbf{SVHN} (source) to \textbf{MNIST} (target). The target features in \protect\subref{r2} and \protect\subref{r3} are in close proximity of the source features but only weakly aligned to them. Source clusters are imperfectly delineated, demonstrating that the learned source features are insufficiently discriminative. Our new discrepancy measure in \protect\subref{r4} leads to improved separation of source features and better alignment between the source and the target, which adheres to the structure of the data in both domains.}
     \label{ee1}
\end{figure*}

Existing discrepancy approaches, reviewed in the next section, mainly focus on aligning domain-level feature distributions without considering category-level alignment. Thus, the alignment enforced by such discrepancy measures does not guarantee a good target performance as it ignores the cluster structure of the samples, aligned with their task labels. The assumption that the source features exhibit a well-defined cluster structure naturally transfers to the target: target features indicative of the same tasks as the source should manifest a similar cluster structure. In other words, when optimally aligned, the target features should amass around the source clusters such that the decision boundaries of the learned task classifiers do not induce partitioning of smooth clusters of target features. However, the aforementioned domain discrepancy measures only focus on global feature overlap, ignoring the finer task-aligned structure in the data. Consequently, they may inaccurately match the clusters and also cause the source features to form weakly separable clusters, as illustrated in \autoref{ee1}, \subref{r2}, \subref{r3}.


To alleviate the limitations of existing discrepancy measures for domain adaptation, we introduce a task (e.g., classification)-specific adversarial discrepancy measure that extends the discriminator output over the source classes, in order to additionally incorporate task knowledge into the adversarial domain alignment. The new discrepancy measure helps the feature extractor (encoder) make discriminative source/target features by considering the decision boundary information. Consequently, source-target alignment not only takes into account the domain-level feature densities but also the category-conditioned clusters-of-features information to produce an improved overlap, evident in \autoref{ee1}, part \subref{r4}.

Motivated by the information-bottleneck principle~\cite{tishby2015deep}, whose goal is to improve generalization by ignoring irrelevant (domain-variant) distractors present in the original data features, we also introduce a \textbf{source regularization} loss by minimizing the information between the source samples and their features by encouraging the marginal distribution of the source features to be similar to a prior distribution (the standard normal distribution) to enforce the model to focus only on the most discriminative (label-variant) features, less prone to overfitting. Moreover, an additional \textbf{target regularization} term is imposed on the classifier, trained on the shared features of the source samples, to encourage it not to pass through high-density regions of the target data. Previous \textbf{DA} methods did not explicitly consider these desiderata. Our ablation study in \autoref{as} empirically demonstrates the importance of the introduced objectives. We also empirically evaluate the advantages of our proposed method by demonstrating considerable improvements over the state-of-the-art methods on several standard domain adaptation benchmarks, including Digits, \textbf{PACS} and \textbf{VisDA} datasets.

\section{Related Work}

We summarize DA works most relevant to this paper. Several types of adversarial learning methods for unsupervised domain adaptation have been shown to match distributions of the features generated from source and target samples~\cite{donahue2013semi,kumar2010co,sun2015return,chen2011co,fernando2013unsupervised,long2016unsupervised}.

The domain adversarial neural network (\textbf{DANN})~\cite{ganin2016domain} first introduced a gradient reversal layer that reversed the gradients of the domain discriminator in order to encourage domain confusion.  Other recent proposals~\cite{NIPS2016_6544,bousmalis2017unsupervised} have explored generative models such as GANs~\cite{goodfellow2014generative} to generate synthetic images for domain adaptation.  These approaches typically train two GANs on the source and target input data with tied parameters with the goal of translating images between the domains. Despite being visually compelling, such image-space models have only been shown to work on small images and limited domain shifts.

In order to circumvent the need to generate images, \textbf{ADDA}~\cite{tzeng2017adversarial} recently proposed an adversarial framework for directly minimizing the distance between the source and target encoded representations (shared features). A discriminator and (target) encoder are iteratively optimized in a two-player game, where the goal of the discriminator is to distinguish the target features from the source features, with the goal of the encoder being to confuse the discriminator. The \textbf{DupGAN}~\cite{hu2018duplex} proposed a \textbf{GAN}-like model with duplex discriminators to restrict the latent representation to be domain invariant, with its category information preserved. Saito et al.~\cite{saito2018maximum} further introduce two classifiers as a discriminator to avoid ambiguous features near the class boundaries. By deploying two classifiers, the method therein employs the adversarial learning techniques to detect the disagreement across classifiers, such that the encoder is able to minimize this discrepancy on target samples. 

In addition to the adversarial distribution matching oriented algorithms, pseudo-labels or conditional entropy regularization are also adopted in literature~\cite{saito2017asymmetric,sener2016learning,zhang2018collaborative}. Sener et al.~\cite{sener2016learning} construct a k-NN graph of target points based on a predefined similarity graph. Pseudo-labels are assigned to target samples via their nearest source neighbors, which allows end-to end joint training of the adaptation loss. Saito et al.~\cite{saito2017asymmetric} employ the asymmetric tri-training, which leverages target samples labeled by the source-trained classifier to learn target discriminative features. Zhang et al.~\cite{zhang2018collaborative} iteratively select pseudo-labeled target samples based on their proposed criterion and retrain the model with a training set including pseudo-labeled samples. However, these methods based on pseudo-labeled target samples have a critical bottleneck where false pseudo-labels can mislead learning of target discriminative features, leading to degraded performance. 

\section{Method}\label{pm}
\subsection{Problem Formulation}\label{pf}
Without loss of generality, we consider a multi-class ($K$-class) classification problem as the running task example. 
{\em Consider the joint space of inputs and class labels, $\mathcal{X}\times\mathcal{Y}$ where 
$\mathcal{Y}=\{1,\dots,K\}$ for ($K$-way) classification. Suppose we have two domains on this joint space, \textbf{source (S)} and \textbf{target (T)}, defined by unknown distributions $P_S({\bf x}, y)$ and $P_T({\bf x}, y)$, respectively.  We are given source-domain training examples with labels $\mathcal{D}_S = \{({\bf x}^S_i, y^S_i)\}_{i=1}^{N_S}$ and target data $\mathcal{D}_T = \{{\bf x}^T_i\}_{i=1}^{N_T}$ with no labels.  We assume the shared set of class labels between the two domains. The goal is to assign the correct class labels $\{y^T_i\}$ to target data points $\mathcal{D}_T$.}

To tackle the problem in the shared latent space framework, we introduce a shared encoder $Q$ between the source and the target domains that maps a sample $\mathbf{x}$ into a stochastic embedding\footnote{Please see Remark~\ref{remark_stochastic} for the benefits of choosing a stochastic encoder over a deterministic one.} $\mathbf{z}\sim Q(\mathbf{z}|\mathbf{x})$, and then apply a classifier $h$ to map $\mathbf{z}$ into the label space $y \sim h(y|\mathbf{z})$ ($h$ is trained
to classify samples drawn from the encoder distribution). Although one can consider domain-wise different encoders, more recent \textbf{DA} approaches tend to adopt a shared encoder, which can prevent domain-specific nuisance features from being learned, reducing potential overfitting issues. 
We define the stochastic encoder $E$ as a \textbf{conditional Gaussian} distribution with \textbf{diagonal} covariance that has the form $Q(\mathbf{z}|\mathbf{x})=\mathcal{N}(\bf z|f_{\mu}(\mathbf{x}),f_{\Sigma}(\mathbf{x}))$ where $\mathbf{f}_\cdot$ is a deep network mapping the data point $\mathbf{x}$ to the $2p$-dimensional latent code, with the first $p$ outputs from $f_e$ encoding $\mathbf{f}_{\mu}$, and the remaining $p$ outputs encoding $\mathbf{f}_{\Sigma}$ (in this work, we set $p=256$ for all the experiments). 
The classifier $h$ outputs a $K$-dim probability vector of class memberships, modeled as a softmax form $\mathbf{h}(\mathbf{z}) = \softmax(\mathbf{f}_c(\mathbf{z}))$,
where $\mathbf{f}_c(\mathbf{z})$ is a deep network mapping the latents $\bf z$ to the logits of $K$ classes.

\begin{remark}\label{remark_stochastic}
The reason to choose a stochastic encoder over a deterministic one is two fold. First, it allows one to impose smoothness (local-Lipschtizness) constraint on the classifier $h$ over target samples; see \autoref{tar} for more details. Second, adding continuous noise to the inputs of the discriminators has been shown to improve instability and vanishing gradients in adversarial optimization problems through smoothening the distribution of features~\cite{Arjovsky2017TowardsPM}. Our stochastic encoder equipped with the reparametarization approach inherently provides such mechanism to feature distribution smoothness; see \autoref{discre1} and \autoref{discre2} for more details.   
\end{remark}

The proposed domain adaptation method can be summarized by the objective function consisting of six terms:
\begin{align}
    \mathcal{L}_{Class} + \mathcal{L}_{Disc} + \mathcal{L}_{Teach} + \mathcal{L}_{Smooth} + \mathcal{L}_{Entropic} + \mathcal{L}_{Adv},
\end{align}
where $\mathcal{L}_{Class}$ is the classification loss applied to $\mathcal{D}_S$, $\mathcal{L}_{Disc}$ is the domain discrepancy loss measuring the discrepancy between the source and target distribution, $\mathcal{L}_{Teach}$ is the source-to-target teaching loss, which couples the source classifier with the target discriminator. The remaining losses,  $\mathcal{L}_{Smooth}, \mathcal{L}_{Entropic}, \mathcal{L}_{Adv}$ will impose different regularization constraints on the model:  $\mathcal{L}_{Smooth}$ will impose Lipschitz classifiers in the target space, $\mathcal{L}_{Entropic}$ will strive to drive the classifier towards regions of low density in the same target space, while $\mathcal{L}_{Adv}$ will impose regularization towards a reference density in the shared space $\mathcal{Z}$. We next discuss each of the above losses in more detail and then propose an algorithm to efficiently optimize the desired objective. 

\subsubsection{Source Classification Loss $\mathcal{L}_{Class}$}
Having access to source labels, the stochastic mappings $Q$ and $h$ are trained on source samples to correctly predict the class label by minimizing the standard cross entropy loss,
\begin{equation}\label{classifier_loss}
    \mathcal{L}_{Class}(Q,h):=-\mathbb{E}_{{\mathbf{x}},y\sim P_S(\mathbf{x},y)}\Big[ \mathbb{E}_{\mathbf{z}\sim Q(\mathbf{z}|\mathbf{x})}\big[ \mathbf{y}^{\top}\log h(\mathbf{z}) \big] \Big],
\end{equation}
where $\mathbf{y}$ is the $K$-D one-hot vector representing the label $y$.

\subsubsection{Domain Discrepancy Loss $\mathcal{L}_{Disc}$}\label{discre2}
Since the stochastic encoder $Q$ is shared between the source and target samples, to make sure the source and the target  features are well aligned in the shared space and respect the cluster structure of the original samples, we propose a novel domain alignment loss, which will be optimized in adversarial manner.

Rather than using the standard adversarial approach to minimizing the alignment loss between the source and the target densities in the shared space $\mathcal{Z}$, i.e., finding the encoder $Q$ which "fools" the best binary discriminator $D$ trying to discern source from target samples, our approach is inspired by semi-supervised
GANs~\cite{dai2017good} where it has been found that incorporating task knowledge into the discriminator can
jointly improve classification performance and quality of images produced by the generator. 
\begin{figure*}[tb]
     \vspace{-1.2em}
     \centering
     \includegraphics[trim={0cm 3cm 0cm 1cm},height=0.4\linewidth, width=0.7\linewidth]{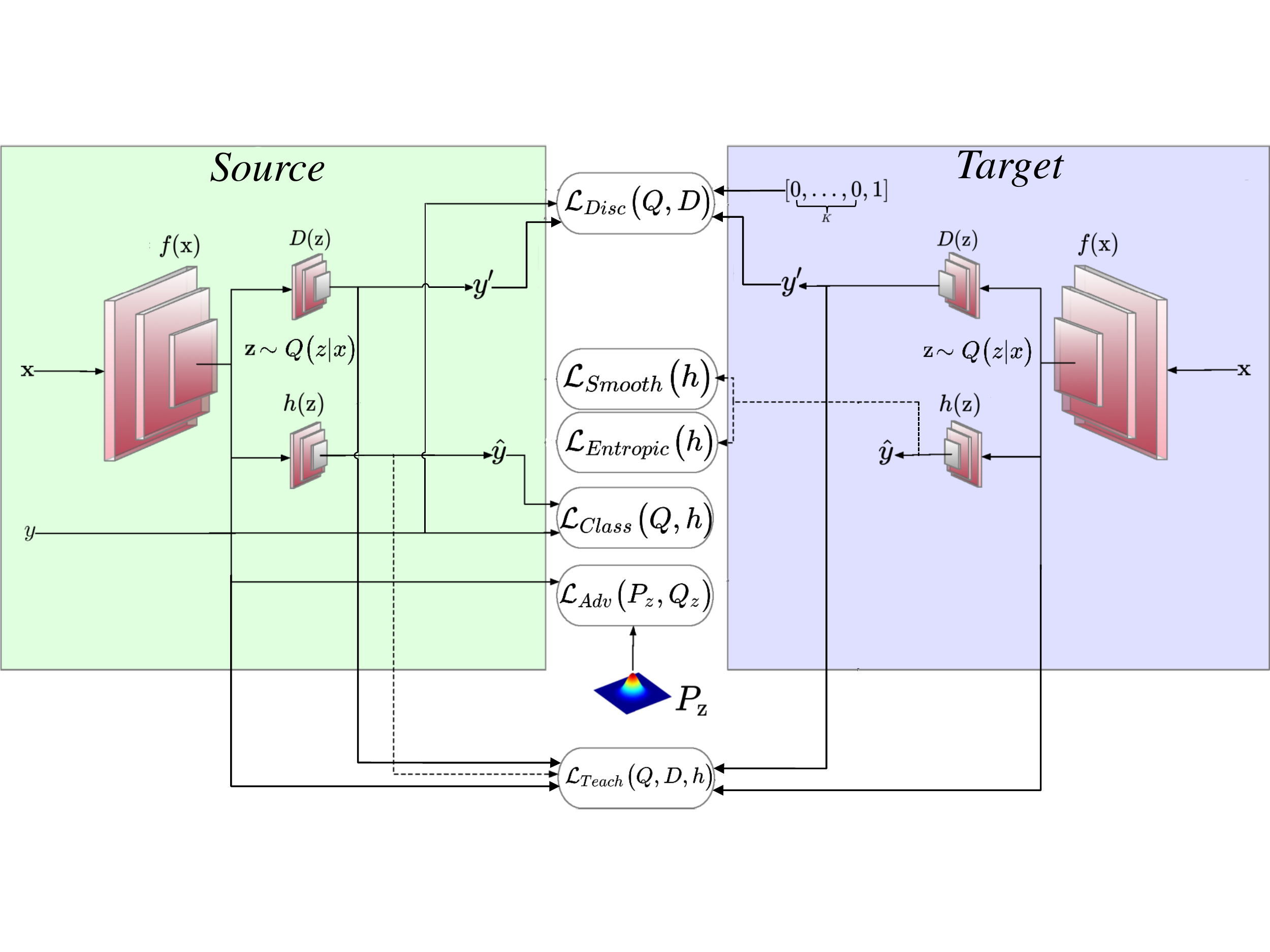}
     \caption{\small 
     \textbf {Proposed architecture} includes a deep feature extractor $f(x)$ and a deep
label predictor $h(z)$, which together form a standard feed-forward architecture.
Unsupervised domain adaptation is achieved by adding a task-specific discriminator $D(z)$ connected to the feature extractor distinguishing the source from target features. The training proceeds standardly and minimizes the label
prediction loss (for source examples) $\mathcal{L}_{Class}$, the domain discrepancy losses (for all samples) $\mathcal{L}_{Disc}$ and $\mathcal{L}_{Teach}$, the source domain regularization loss $\mathcal{L}_{Adv}$, and the target domain regularization losses $\mathcal{L}_{Smooth}$ and $\mathcal{L}_{Entropic}$.
     }
     \label{model}
     \vspace{-1em}
\end{figure*}
We incorporate task knowledge by replacing what would be a binary discriminator with a $(K+1)$-way multi-class discriminator $y' = D(\mathbf{z}) = \softmax(\mathbf{f}_d(\mathbf{z}))$.
The first $K$ classes indicate that a sample $\mathbf{z}$ belongs to the source domain {\em and} belongs to a specific classes in $\mathcal{Y}$, while the last $(K+1)$-th class $"t"$ indicates $\mathbf {z}$ belongs to the target domain.

 Since we have the class label for the source samples, the discriminator is trained to classify source features correctly, hence creating crisp source clusters in the feature space. On the other hand, the new discriminator seeks to distinguish the samples from the target domain from those of the source by assigning them to the $(K+1)$-th, "target" class.
\begin{multline}\label{task_specific_loss}
    \mathcal{L}_{Disc}(Q,D) := 
    -\mathbb{E}_{\mathbf {x}\sim P_T(\mathbf {x})}\big[ \mathbb{E}_{\mathbf {z}\sim Q(\mathbf {z}|\mathbf {x})}\big[ [\mathbf{0},1]^{\top}\log D(\mathbf {z}) \big] \big] \\
    - \mathbb{E}_{\mathbf {x},y\sim P_S(\mathbf {x},y)}\big[ \mathbb{E}_{\mathbf {z}\sim Q(\mathbf {z}|\mathbf {x})}\big[ [\mathbf{y},0]^{\top}\log D(\mathbf {z}) \big]\big], 
\end{multline}
where $[\mathbf{0},1]$ is a one-hot vector indicating a point from the target domain and $[\mathbf {y},0]$ stands for a point from the source domain, labeled according to class label $\mathbf{y}$. 
\subsubsection{Teacher Target-Source Loss $\mathcal{L}_{Teach}$}
Here, we seek the encoder $Q$ to generate a feature representative of one of the first $K$ task-specific classes for target samples preserving their cluster structure and aligning them to the source clusters in the feature space. However, the target data points are unlabeled, and the encoder will not have the chance to enforce the desired clustering structure of the target points, where points within a cluster would have the same predicted label. To "teach" the encoder, we ask the classifier $h(\cdot)$ to provide pseudo soft labels for the target points to our new discriminator using the following loss:
\begin{multline}\label{teach_loss}
    \mathcal{L}_{Teach}(Q,D,h):= \\
    -\mathbb{E}_{\mathbf {x}\sim P_T(\mathbf {x})}\bigg[ \mathbb{E}_{\mathbf {z}\sim Q(\mathbf {z}|\mathbf {x})}[ [h(\mathbf {z}),0]^{\top}\log D(\mathbf {z})] \bigg].
\end{multline}
Intuitively, the encoder tries to fool the discriminator by assigning one of the first $K$ classes to target features, leveraging on the output of the classifier $h$ (augmented with $0$ for the $K+1$-th dimension) as pseudo-labels for target features.
\begin{remark}
The proposed task-specific domain discriminator can be used to improve any domain adaptation method that has an adversarial domain alignment component. Indeed, we observe (see \autoref{ex:task}) that the proposed discriminator significantly improves upon the standard binary discriminator.
\end{remark}

\subsubsection{Source Domain Regularization Loss}\label{discre1}

One of the standard goals in representation learning is to find an encoding of the data point $\mathbf{x}$ that is maximally expressive about its label $y$ while being maximally compressive about $\mathbf{x}$---finding a representation $\mathbf{z}$ which ignores as many details of $\mathbf{x}$ as possible. This is specifically useful for domain adaptation where we require a representation to be domain invariant. Essentially, we want $\mathbf {z}$ to act like a minimal sufficient statistic of $\mathbf {x}$ for predicting $y$ in order to generalize better for samples form unseen domains. To do so, we introduce a regularizer that acts on the aggregated posterior of the shared features of the source samples $Q_z(\mathbf{z}) = \mathbb{E}_{\mathbf{x} \sim P_S(\mathbf{x})} [ Q(\mathbf{z}| \mathbf{x})]$. The regularizer encourages $\mathbf {z}$ to be less informative about $\mathbf {x}$ in the form of mutual information by matching the aggregated posterior of the shared features with a factorized prior distribution $P_z(\mathbf {z})$\footnote{In this work, we consider $P_z(\mathbf {z}) = \mathcal{N}(0,\mathbf {I})$}, which in turn constrains the implicit capacity of $\mathbf {z}$ and encourages it be factorized:
\begin{equation}
     \mathcal{D} \big[P_z(\mathbf {z})||Q_z(\mathbf {z})\big],
\end{equation}
where $\mathcal{D}(\cdot||\cdot)$ is an arbitrary distribution divergence measure. 

As the proxy for this divergence, we define an auxiliary loss which will be \textbf{adversarially} optimized. We introduce an a binary discriminator $F$ in the latent space trying to separate \textbf{\textit{true}} points sampled from $P_z$ and \textbf{\textit{fake}} ones sampled from $Q_z$. The encoder $Q$ ensures the aggregated posterior distribution $Q_z$ can fool the binary discriminator into thinking that the source features comes from the distribution $P_z$:
\begin{multline}\label{adaptation_loss}
 \mathcal{L}_{Adv}(Q,F) = -\mathbb{E}_{x\sim P_S(\mathbf {x})}\big[ \mathbb{E}_{\mathbf{z}\sim Q(\mathbf {z}|\mathbf {x})}\big[ \log F(\mathbf {z}) \big]\big] \\
    -\mathbb{E}_{\mathbf{z}\sim P(\mathbf {z})}\big[ \log (1-F(\mathbf {z})) \big].
\end{multline}
\begin{remark}
We empirically observed that imposing such regularization on target samples could be harmful to performance. We conjecture this is due to the lack of true class labels for the target samples, without which the encoder would not preserve the label information of the features, leading to unstructured target points in feature space. 
\end{remark}


\subsubsection{Target Domain Regularization Losses}\label{tar}
In order to incorporate the target domain information into the model, we apply the cluster assumption, which states that the target data points $\mathcal{D}_T$ contains clusters and that points in the same cluster have homogeneous class labels. If the cluster assumption holds, the optimal decision boundaries should occur far away from data-dense regions in the feature space $z$. We achieve this by defining an entropic loss,
\begin{multline}\label{target1_loss}
     \mathcal{L}_{Entropic}(h,Q):= \\ 
     -\mathbb{E}_{\mathbf {x}\sim P_T(\mathbf {x})} \bigg[ \mathbb{E}_{\mathbf {z}\sim Q(\mathbf {z}|\mathbf {x})}\big[ h(\mathbf {z})^{\top} \log h(\mathbf {z}) \big] \bigg].
\end{multline}
Intuitively, minimizing the conditional entropy forces the classifier to be confident on the unlabeled target data, thus driving the classifier’s decision boundaries away from the target data. In practice, the conditional entropy must be empirically estimated using the available data.
 
However, Grandvale~\cite{grandvalet2005semi} suggested this approximation can be very poor if $h$ is not locally-Lipschitz smooth. Without the smoothness constraint, the classifier could abruptly change its prediction in the neighborhood of training samples, allowing decision boundaries close to the training samples even when the empirical conditional entropy is minimized. To prevent this, we take advantage of our stochastic encoder and propose to explicitly incorporate the locally-Lipschitz constraint in the objective function,
\begin{multline}\label{target2_loss}
     \mathcal{L}_{Smooth}(h,Q):= \\ \mathbb{E}_{\mathbf {x}\sim P_T(\mathbf {x})} \bigg[ \mathbb{E}_{\mathbf{z}_1,\mathbf{z}_2\sim Q(\mathbf{z}|\bf x)}||h(\mathbf{z}_1) - h(\mathbf{z}_2)||_1 \bigg],  
\end{multline}
with $\|\cdot\|_1$ the $L_1$ norm. Intuitively, we enforce classifier consistency over proximal features of any target point $\mathbf{x}$. 
\begin{remark}
We empirically observed that having such constraints for source features would not improve performance. This is because access to the source labels and forcing the classifier to assign each source feature to its own class would already fulfill the smoothness and entropy constraints on the classifier for the source samples. 
\end{remark}

\subsection{Model Learning and Loss Optimization}


Our goal, as outlined in \autoref{pf}, is to train the task-specific discriminator $D$, binary discriminator $F$, classifier $h$, and encoder $Q$ to facilitate learning of the cross-domain classifier $h$. By approximating the expectations with the sample averages, using the stochastic gradient Descent (SGD), and the reparametarization approach~\cite{kingma2013auto}, we solve the optimization task in the following four subtasks. The overall algorithm is available in the Supplementary Material (\textbf{SM}).

\vspace{-1.4em}\subsubsection{Optimizing the encoder $Q$}
\begin{multline}\label{op_encoder}
  Q^* = \arg \min_Q \; \mathcal{L}_{Class}(Q,h^*) + \mathcal{L}_{Disc}(Q,D^*,h^*) \\
  + \lambda_{Q} \big[\mathcal{L}_{Adv}(Q,F^*)\big],
\end{multline}
where $\lambda_{Q}$ is a weighting factor. Intuitively, The first term in \autoref{op_encoder} encourages $Q$ to produce discriminative features for the labeled source samples to be correctly classified by the classifier $h$. The second term simulates the adversarial training by encouraging $Q$ to fool the task-specific discriminator $D$ by pushing the target features toward the source features, leveraging the soft pseudo-labels provided by the classifier. Through the last term, the encoder seeks to fool the binary discriminator $F$ into treating the source features as if they come from the fully-factorized $P(\mathbf z)$ to produce domain-invariant source features. 

\subsubsection{Optimizing the classifier $h$}
\begin{multline}\label{op:classifier}
    h^* = \arg \min_h\;\lambda_h[\mathcal{L}_{Class}(Q^*,h)] \\
    + \lambda'_{h}\big[\mathcal{L}_{Entropic}(Q^*,h)
    + \mathcal{L}_{Smooth}(Q^*,h)\big], 
\end{multline}
where $\lambda_{h}$ and $\lambda'_{h}$ are the trade-off factors. Intuitively, we enforce the classifier $h$ to correctly predict the class labels of the source samples by the first term in \autoref{op:classifier}. We use the second term to minimize the entropy of $h$ for the target samples, reducing the effects of "confusing" labels of target samples. The last term guides the classifier to be locally consistent, shifting the decision boundaries away from target data-dense regions in the feature space.    

\subsubsection{Optimizing the task-specific discriminator $D$}
\begin{align}\label{op:task_discriminator}
    D^* = \arg \min_D\;\mathcal{L}_{Disc}(Q^*,D).
\end{align}
The loss in \autoref{op:task_discriminator} prompts $D$ to shape its decision boundary to separate the source features (according to their class label) and target features from each other.

\subsubsection{Optimizing the binary discriminator $F$}
\begin{align}\label{op:binary_discriminator}
    F^* = \arg \min_F\;\mathcal{L}_{Adv}(Q^*,F).
\end{align}
Intuitively, the loss in \autoref{op:binary_discriminator} encourages $F$ to separate the source features from the features generated from the fully-factorized distribution $P_z(\mathbf {z})$ by assigning label $1$ and $0$ to the source feature samples and $P_z(\mathbf {z})$ samples, respectively.

\subsection{Target Class Label Prediction}
After model training, to determine the target class-label $y_t$
of a given target domain instance $\mathbf{x}_t$, we first compute the distribution of $y_t$ given $\mathbf{x}_t$ by integrating out the shared feature $\mathbf{z}_t$. Then, we select the most likely label as
\begin{equation}
\hat{y}_t = {\arg \max}_{y_t\in \{1,...,K\}}\; P(y_t|\mathbf{x}_t),
\end{equation}
where $P(y_t|\mathbf{x}_t)$ can be computed as
\begin{align}\label{target_pre}
P(y_t = k|\mathbf{z}_t)=\mathbb{E}_{\mathbf z \sim Q(\mathbf{ z}|\mathbf{x}_t)=\mathcal{N}(f_e^{\mu}(\mathbf{x}_t),f_e^{\Sigma}(\mathbf{x}_t))}[ h_k(\mathbf {z})],
\end{align}
where $h_k(.)$ is the $k$-th entry of the classifier output.
Since the above expression cannot be computed in a closed form, we approximate it with its mean value. Using this approximation, we compute $y_t$ as:
\begin{equation}
\hat{y}_t = {\arg \max}_{k\in \{1,...,K\}}\; h_k(\mathbf{z}_t),\;\;\;\mathbf{z}_t = f_e^{\mu}(\mathbf{x}_t).
\end{equation} 
\begin{remark}
We empirically observed that estimating the expectation in \autoref{target_pre} with Gibbs sampling from the posterior $Q(\mathbf{z}|\mathbf{x}_t)$ instead of its mean would not boost the performance. We conjecture this is due to the smoothness constraint we impose on the classifier through \autoref{target2_loss}, enforcing consistency over proximal target samples drawn from $Q(\mathbf{z}|\mathbf{x})$.  
\end{remark}
\section{Experimental Results}\label{er}
\begin{figure}[t]
 \centering
   
  \begin{subfigure}[t]{.30\linewidth}
    \includegraphics[trim = 5mm 2mm 0mm 0mm, clip, scale=0.142]{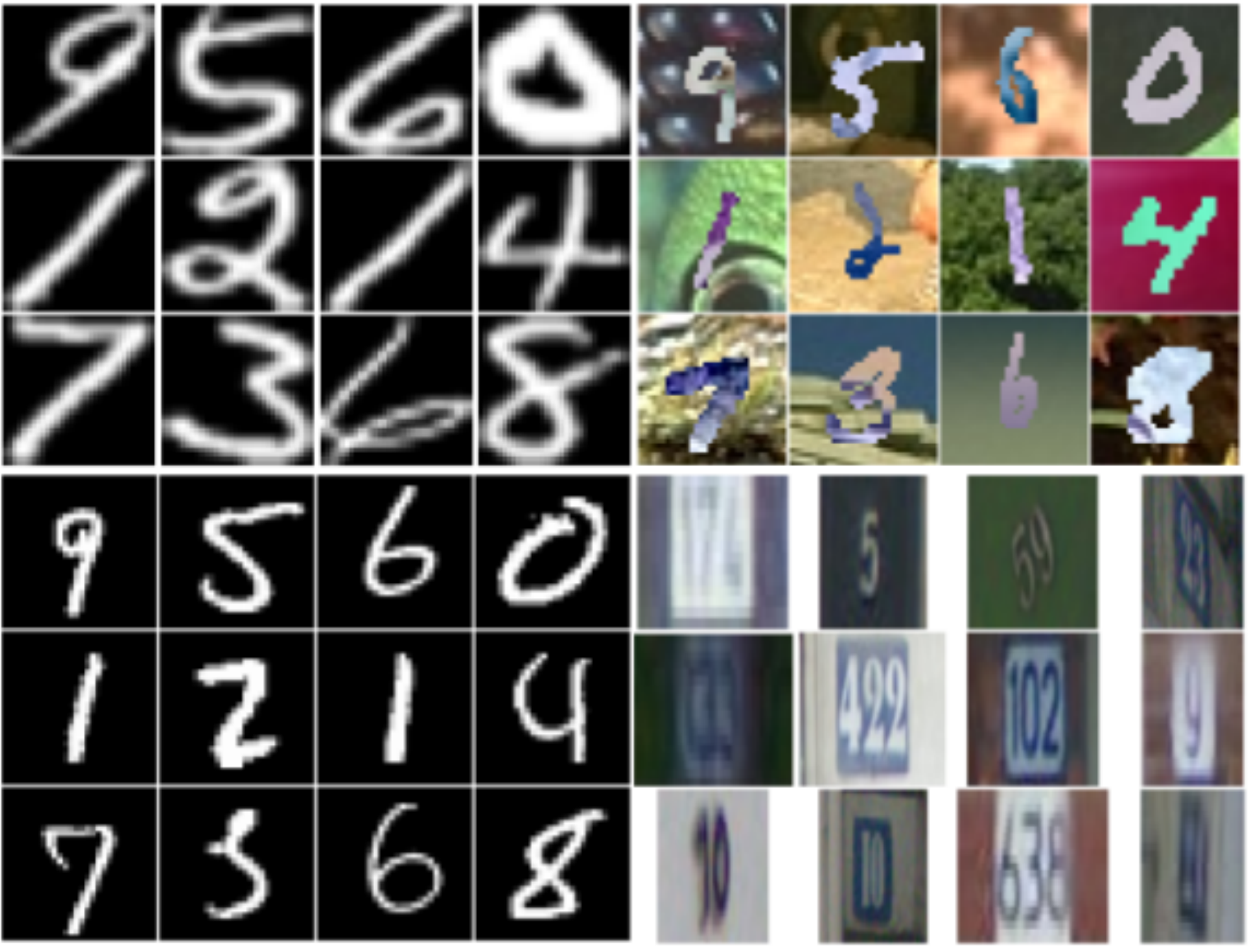} 
    \caption{Digits.}
 \end{subfigure} \
 \begin{subfigure}[t]{.30\linewidth}
    \includegraphics[trim = -15mm 2mm 0mm 0mm, clip, scale=0.118]{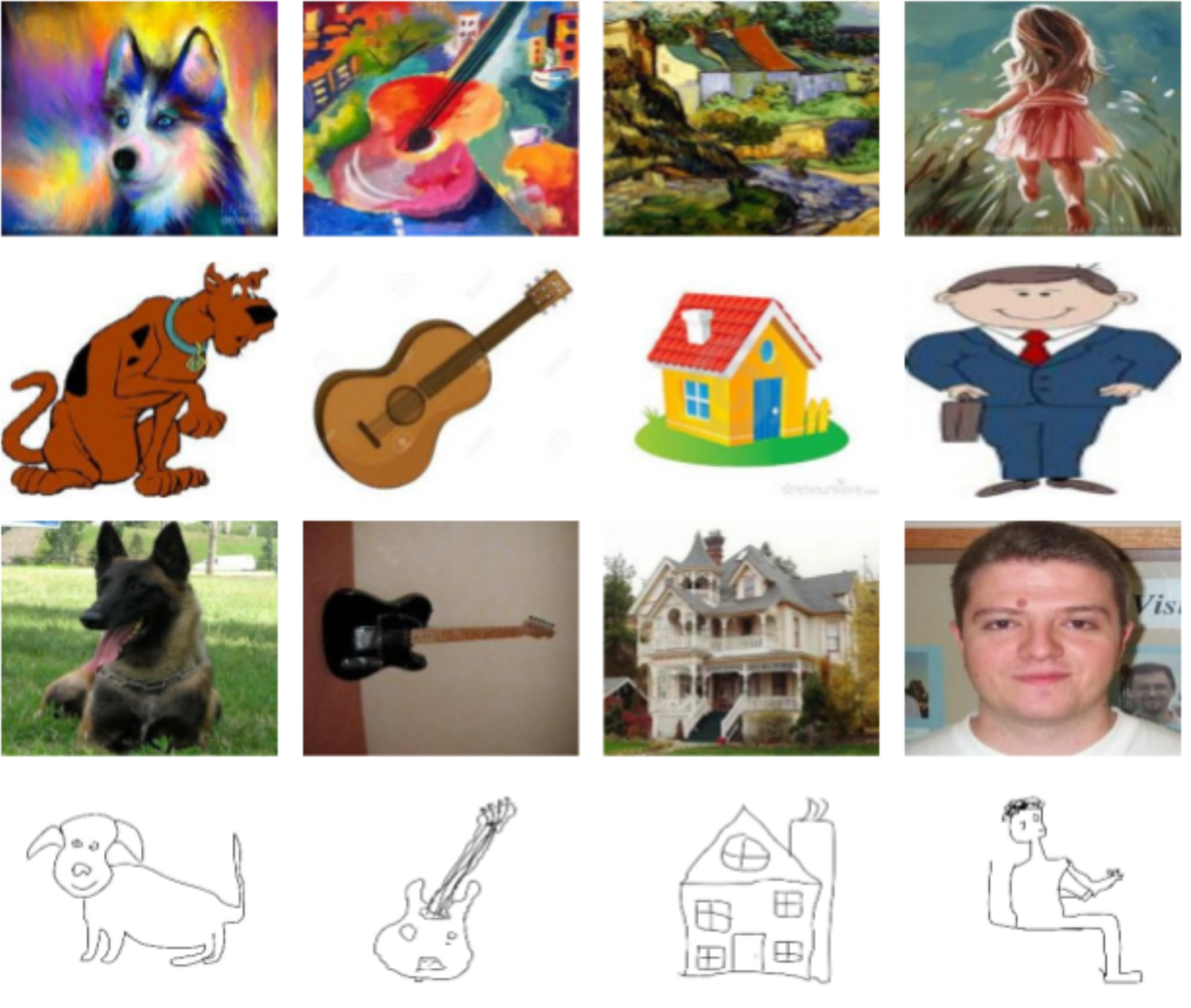}
    \caption{PACS}
  \end{subfigure} \
  \begin{subfigure}[t]{0.30\linewidth}
    \includegraphics[trim = 5mm 2mm 5mm 0mm, clip, scale=0.070]{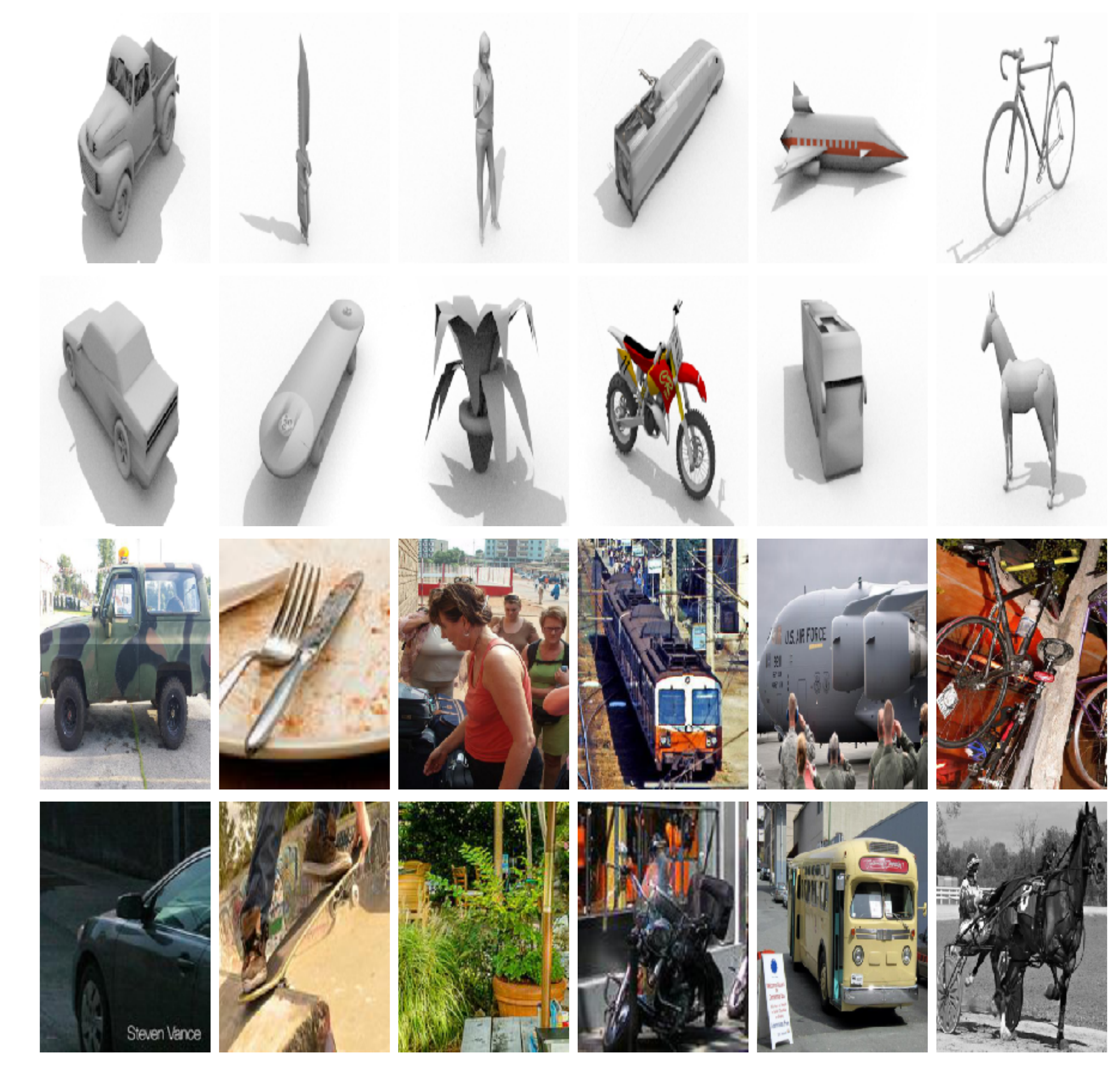}
    \caption{VisDA.}
  \end{subfigure}
\vspace{-0.8em}
\caption{Example images from benchmark datasets. 
}
\label{samples}
\vspace{-0.8em}
\end{figure}

We compare our proposed method with state-of-the-art on three benchmark tasks. The Digit datasets embody the digit cross-domain classification task across four datasets: \textbf{MNIST}, \textbf{MNIST-M}, \textbf{SVHN}, \textbf{USPS}, which consist of $K = 10$ digit classes (0-9). We also evaluated our method on \textbf{VisDA} object classification dataset~\cite{peng2017visda} with more than 280K images across twelve categories. Finally, we report performance on \textbf{PACS}~\cite{li2017deeper}, a recently proposed benchmark which is especially interesting due to the significant domain shift between different domains. It contains images of seven categories extracted from four different domains:\textit{Photo} (P), \textit{Art paintings} (A), \textit{Cartoon} (C), and \textit{Sketch} (S). The details of the datasets are available in \textbf{SM}.  \autoref{samples} illustrates image samples from different datasets and domains. We evaluate the performance of all methods with the classification accuracy metric. We used ADAM~\cite{kingma2014adam} for training; the learning rate was set to $0.0002$ and momentums to $0.5$ and $0.999$. Batch size was set to $16$ for each domain, and the input images were mean-centered/rescaled to $[-1,1]$. All the used architectures replicate those of state-of-the-art methods, detailed in \textbf{SM}. We followed the protocol of unsupervised domain adaptation and did not use validation set to tune the hyper-parameters $\lambda_{Q},\lambda_{h},\lambda'_{h}$. Full hyper-parameter details for each experiment can be found in \textbf{SM}. We compare the proposed method with several related methods, including \textbf{CORAL}~\cite{sun2016deep}, \textbf{DANN}~\cite{ganin2014unsupervised}, \textbf{ADDA}~\cite{tzeng2017adversarial}, \textbf{DTN}~\cite{zhang2015deep}, \textbf{UNIT}~\cite{liu2017unsupervised}, \textbf{PixelDA}~\cite{bousmalis2017unsupervised}, \textbf{DIAL}~\cite{carlucci2017just}, \textbf{DLD}~\cite{mancini2018boosting}, \textbf{DSN}~\cite{bousmalis2016domain}, and \textbf{MCDA}~\cite{saito2018maximum} on digit classification task (Digit datasets), and the object recognition task (VisDA and PACS datasets).

\subsection{Results On Digits Recognition}
In this evaluation, we follow the same protocol across all methods. Specifically, 
we use the network structure similar to \textbf{UNIT}~\cite{liu2017unsupervised}. See SM for more details.

We show the accuracy of different methods (averaged over five different runs) in \autoref{1}. The proposed method outperformed the competing methods in five out of six  settings, confirming consistently and significantly better generalization of our model over target data. 

The higher performance of the proposed model compared to other methods is mainly attributed to the proposed task-specific alignment method, which not only encourages the source features to be well-separated, according to their class label, but also aligns the target to source features in a cluster-wise manner, "matching" the source and target clusters. This is in contrast to the standard domain-wise alignment, which ignores the source/target inherent cluster structure. This superiority also benefits from the proposed source and target domain regularizers, which improve the source feature domain-invariance and the classifier's robustness respectively. See \autoref{as} for more details. 
\begin{table}[t]
	\caption{Mean classification accuracy on digit classification. M: MNIST; MM: MNIST-M, S: SVHN, U: USPS. The best is shown in red. The superscript shows the standard deviation. *\textbf{UNIT} trains with the extended \textbf{SVHN} ($>500$K images vs ours $72$K). *\textbf{PixelDA} uses ($\approx 1,000$) of labeled target domain data as a validation set for tuning the hyper-parameters.}
	\centering
	\resizebox{1\columnwidth}{!}{%
		\begin{tabular}[b]{l||l|l|l|l|l|l}
			\hline
			method & S $\rightarrow$ M & M $\rightarrow$ MM& M $\rightarrow$ U & MM $\rightarrow$  M&MM $\rightarrow$ U &U $\rightarrow$ M \\
			\hline
			\textbf{Source Only} &62.10   &55.98 &78.30 &84.46 &80.43  &50.64 \\
			\textbf{1-NN} & 35.86&  12.58& 41.22&  82.13& 36.90  &38.45 \\
			\textbf{CORAL}~\cite{sun2016deep}  & $63.10^{\small{0.8}}$ &$57.70^{\small{0.7}}$ &$81.05^{\small{0.6}}$  &  84.90& 87.54&$85.01^{\small{0.5}}$ \\
			\textbf{DANN}\cite{ganin2016domain} &$73.80^{\small{0.6}}$    &77.40 &$81.60^{\small{0.4}}$  &61.05 &85.34 &$77.40^{\small{0.4}}$   \\
			\hline
			\textbf{ADDA}\cite{tzeng2017adversarial} & $77.68^{\small{1.5}}$ &$91.47^{\small{0.6}}$ &$90.51^{\small{0.3}}$&  $92.82^{\small{0.6}}$&$80.70^{\small{0.6}}$   &$90.10^{\small{0.8}}$  \\
			\hline
			\textbf{DTN}\cite{zhang2015deep} & $81.40^{\small{0.6}}$   &$85.70^{\small{0.4}}$ &$85.80^{\small{0.4}}$  &$88.80^{\small{0.5}}$ &$90.68^{\small{0.4}}$  &$89.04^{\small{0.3}}$ \\
			\hline
			\textbf{PixelDA}\cite{bousmalis2017unsupervised} &--   &\textbf{\color{red}{98.10}}$^*$ &94.10$^*$ &--&--&-- \\
            \textbf{UNIT}\cite{liu2017unsupervised} &90.6$^*$ &--  &92.90 &--  & &90.60 \\
            \hline
			\textbf{DSN}\cite{bousmalis2016domain} & $82.70^{\small{0.3}}$ &$83.20^{\small{0.4}}$ &$91.65^{\small{0.3}}$ & $90.20^{\small{0.3}}$ &$89.95^{\small{0.2}}$ &$91.40^{\small{0.3}}$  \\
			\hline
			\textbf{MCDA}\cite{bousmalis2016domain} & $\textbf{\color{red}{96.20}}^{ \small{0.4}}$ &-- &$96.50^{\small{0.7}}$ & -- &-- &$94.10^{ \small{0.3}}$
			\\
			\hline
			\textbf{Ours} & $94.67^{\small{0.5}}$ &$\textbf{\color{red}{98.01}}^{\small{0.3}}$ &$\textbf{\color{red}{99.05}}^{\small{0.3}}$ &$\textbf{\color{red}{99.11}}^{\small{0.2}}$ &$\textbf{\color{red}{99.16}}^{\small{0.3}}$ &$\textbf{\color{red}{97.85}}^{\small{0.3}}$ 			
		\end{tabular}
	}
	\label{1}
	\vspace{-1em}
\end{table}
\begin{table*}[t]
\begin{center}
\vspace{-1em}
\caption{Accuracy of ResNet101 model fine-tuned on the VisDA dataset. Last column shows the average rank of each method over all classes. The best in bold red, second best in red.}
\label{tb:visda}
\scalebox{0.85}{
\begin{tabular}{l||cccccccccccc|c|c}
Method     & plane & bcycl & bus  & car  & horse & knife & mcycl & person & plant & sktbrd & train & truck & mean & Ave. ranking \\\hline
Source Only & 55.1      & 53.3    & 61.9 & 59.1 & 80.6  & 17.9  & 79.7        & 31.2   & 81.0    & 26.5       & 73.5  & 8.5   & 52.4 & 4.91\\\hline
\textbf{MMD}~\cite{long2015learning}      & \color{red}{87.1}      & 63.0      & 76.5 & 42.0   & {\bf\color{red}{90.3}}  & 42.9  &  {\bf\color{red}{85.9}}        & 53.1   & 49.7  & 36.3       & {\bf \color{red}{85.8}}  & \color{red}{20.7}  & 61.13 & 3.08\\
\textbf{DANN}~\cite{ganin2014unsupervised}       & 81.9      &  \color{red}{77.7}    & \color{red}{82.8} & 44.3 & 81.2  & 29.5  & 65.1        & 28.6   & 51.9  & { \bf\color{red}{54.6}}       & 82.8  & 7.8   & 57.42& 3.00\\
\textbf{MCDA}~\cite{saito2018maximum} & 87.0&60.9&{\bf \color{red}{83.7}}&\color{red}{64.0}&88.9& \color{red}{79.6}&84.7&{\bf\color{red}{76.9}}& \color{red}{88.6}&40.3&\color{red}{83.0}&{\bf \color{red}{25.8}}&{ \color{red}{71.90}} & \color{red}{2.41}\\
\hline\hline
\textbf{Ours}& {\bf \color{red}{88.2}}&{\bf \color{red}{78.5}}& 79.7&{\bf \color{red}{71.1}}&\color{red}{90.0}&{\bf \color{red}{81.6}}&\color{red}{84.9}&\color{red}{72.3}&{\bf \color{red}{92.0}}&\color{red}{52.6}&82.9&18.4&{ \bf\color{red}{74.03}} & {\bf\color{red}{1.83}}
\end{tabular}}
\end{center}
  \vspace{-.5em}

\end{table*}

\subsection{Results on Object Recognition}
We also evaluate our method on two object recognition benchmark datasets \textbf{VisDA}~\cite{peng2017visda} and \textbf{PACS}~\cite{li2017deeper}. We follow \textbf{MCDA}~\cite{saito2018maximum}, and use ResNet101~\cite{he2016deep} as the backbone network which was pretrained on ImageNet dataset, and then finetune the parameters of ResNet101 with the source only \textbf{VisDA} dataset according to the procedure described
in~\cite{saito2018maximum}. For the \textbf{PACS} dataset, we also follow the experimental protocol in~\cite{mancini2018boosting}, using ResNet18~\cite{he2016deep} pretrained on ImageNet dataset, and training our model considering 3 domains as sources and the remaining as target, using all the images of each domain. 
For these experiments, we set the learning rate of resnets to $10^{-9}$. We choose this small learning
rate for ResNet parameters since the domain shift for
both \textbf{VisDA} and \textbf{PACS} are significant, the training procedure benefits from a mild parameter updates back-propagated from the
loss.
\begin{figure}[t]
\vspace{-1.0em}
\begin{center}
\includegraphics[trim = 0mm 5mm 0mm 0mm,height=.4\linewidth, clip]{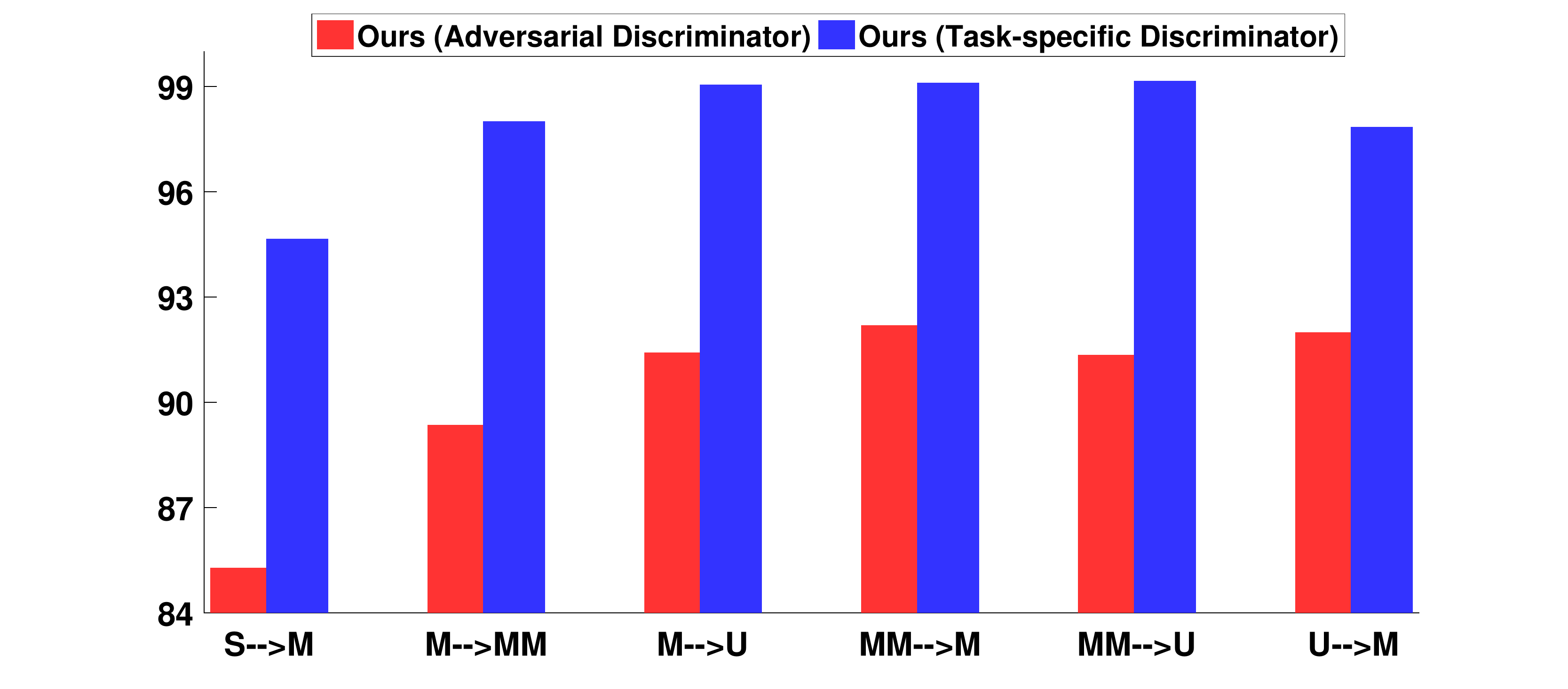}
\end{center}
\vspace{-2em}
\caption{Comparison of proposed task-specific discriminator with the standard adversarial discriminator on Digit dataset.}
\label{ablation2}
\vspace{-1em}
\end{figure}
Results for this experiment are summarized in \autoref{tb:visda} $\&$ \autoref{pacs_re}. We observe that our model achieved, on average, the best performance compared to other competing methods for both datasets. The higher performance of our method is mainly attributed to incorporating the category-level information into the domain alignment through the proposed task-specific discriminator, which is beneficial to boost the discriminability of the source/target features.

\begin{table}[b]
\vspace{-1.0em}
	\caption{Mean classification accuracy on \textbf{PACS} dataset. The first row indicates the target domain, while all the others are considered as sources. The best (in bold red), the second best (in red).}
	\centering
	\resizebox{\columnwidth}{!}{%
		\begin{tabular}[b]{c||c|c|c|c|c}
			\hline
			method & Sketch & Photo & Art & Cartoon & Mean \\
			\hline
			\textbf{Resnet18 (Source Only)} &60.10   &92.90 &74.70 &72.40 &75.00   \\
			\textbf{DIAL}~\cite{carlucci2017just} & 66.80&  \textbf{\color{red}{97.00}}& 87.30&  85.50& 84.20   \\
			\textbf{DLD}~\cite{mancini2018boosting}  & \color{red}{69.60} &\textbf{\color{red}{97.00}} &\color{red}{87.70}  &  \color{red}{86.90}& \color{red}{85.30}
			\\
			\hline
			\textbf{Ours} & \textbf{\color{red}{71.69}}  &\color{red}{96.81} &\textbf{\color{red}{89.48}} &\textbf{\color{red}{88.91}} &\textbf{\color{red}{86.72}}    
		\end{tabular}
	}
	\label{pacs_re}
\end{table}

\begin{figure*}[h]
\vspace{-1.3em}
\centering
     \begin{tabular}{ccc}
    \begin{subfigure}[b]{0.35\linewidth}
    \includegraphics[trim = 0mm 0mm 0mm 0mm,clip,width = .7\linewidth,height = .55\linewidth]{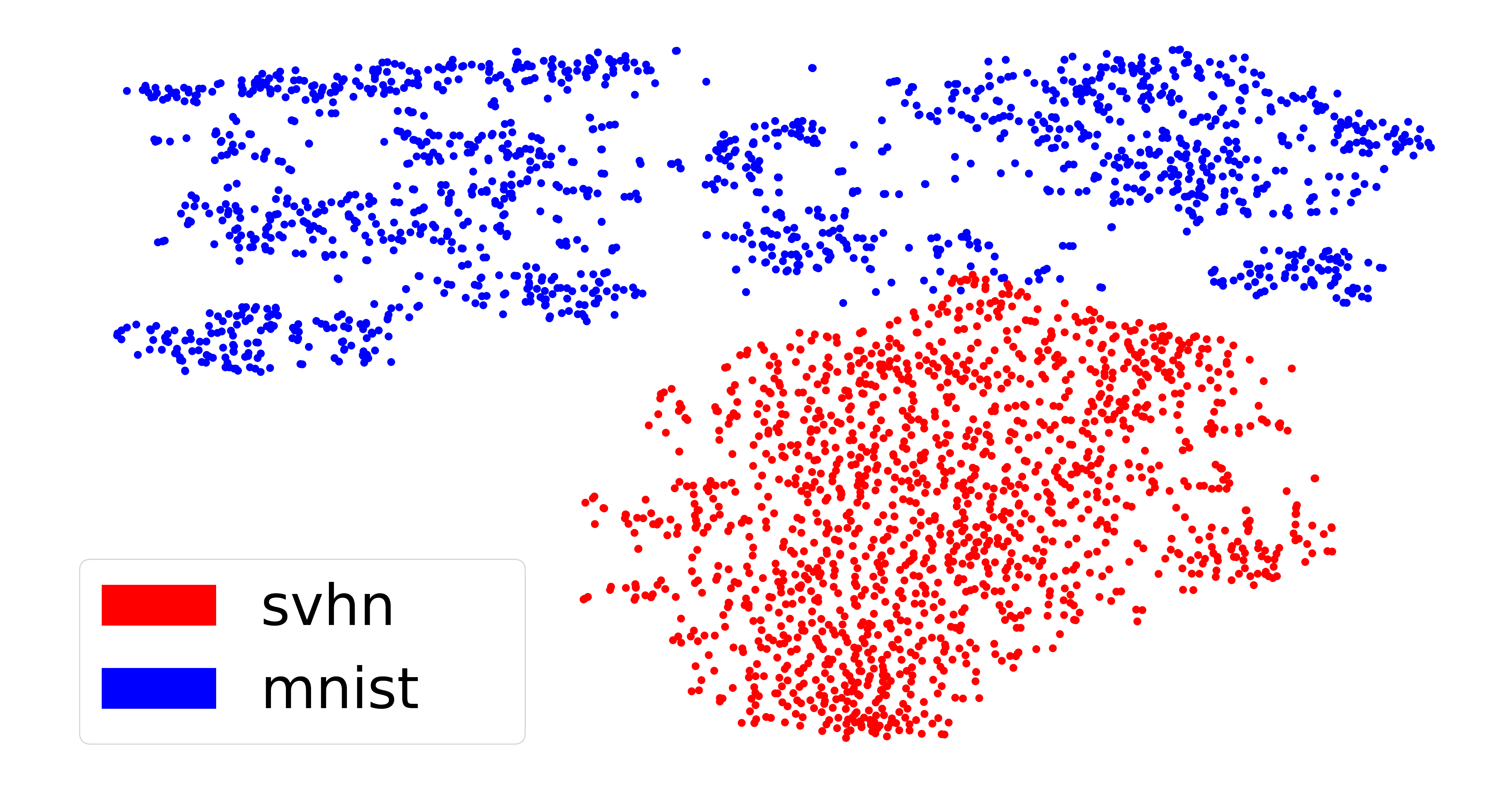}
    \caption{Original (by domain)\label{vis_original}}
    \end{subfigure}
    &
    \begin{subfigure}[b]{0.35\linewidth}
	 \includegraphics[clip,width = .7\linewidth,height = .55\linewidth]{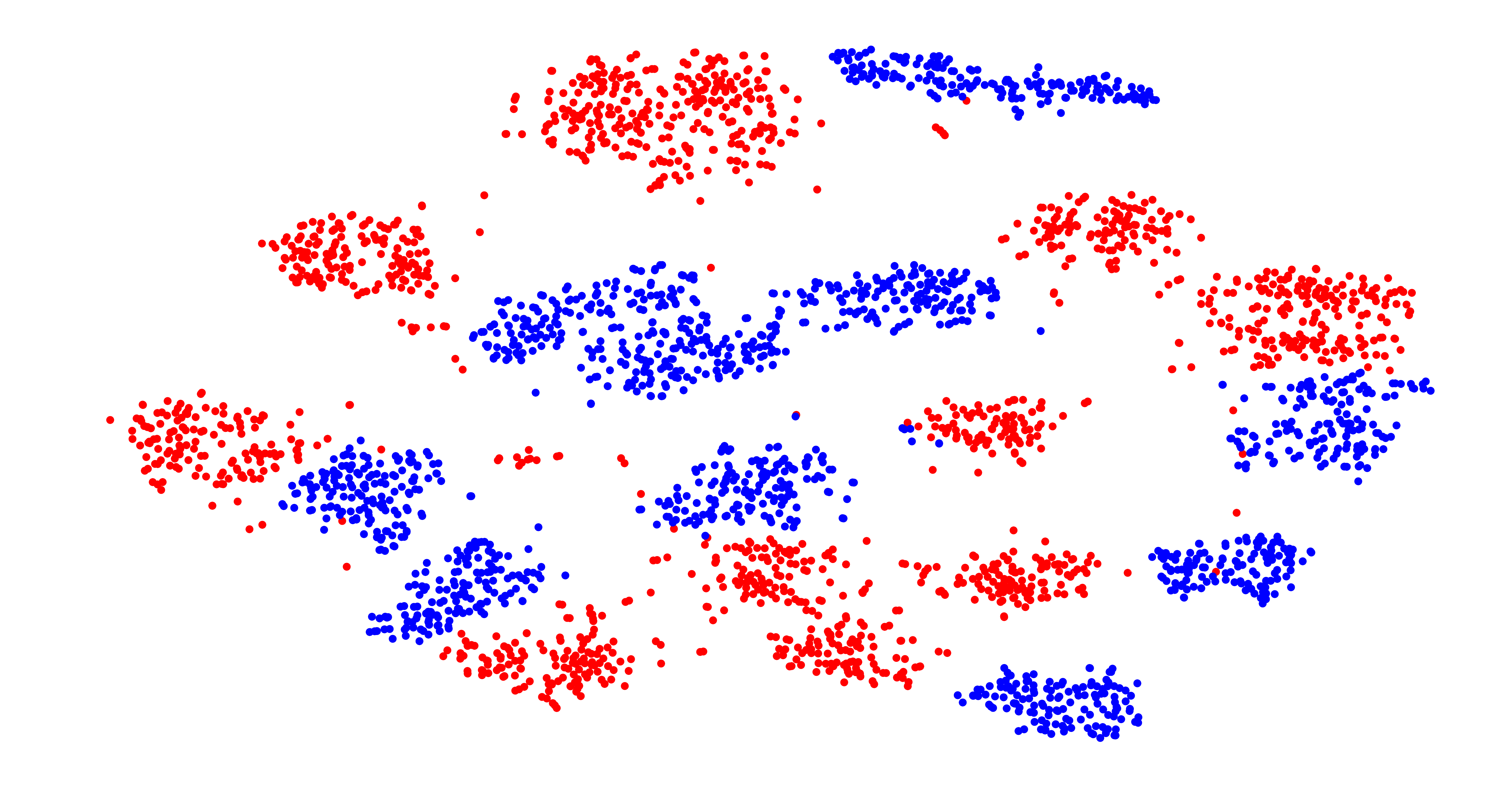}
	 \caption{Adversarial discriminator
	 \label{vis_binary}}
	 \end{subfigure}
	 &
	 \begin{subfigure}[b]{0.35\linewidth}
	 \includegraphics[clip,width = .7\linewidth,height = .55\linewidth]{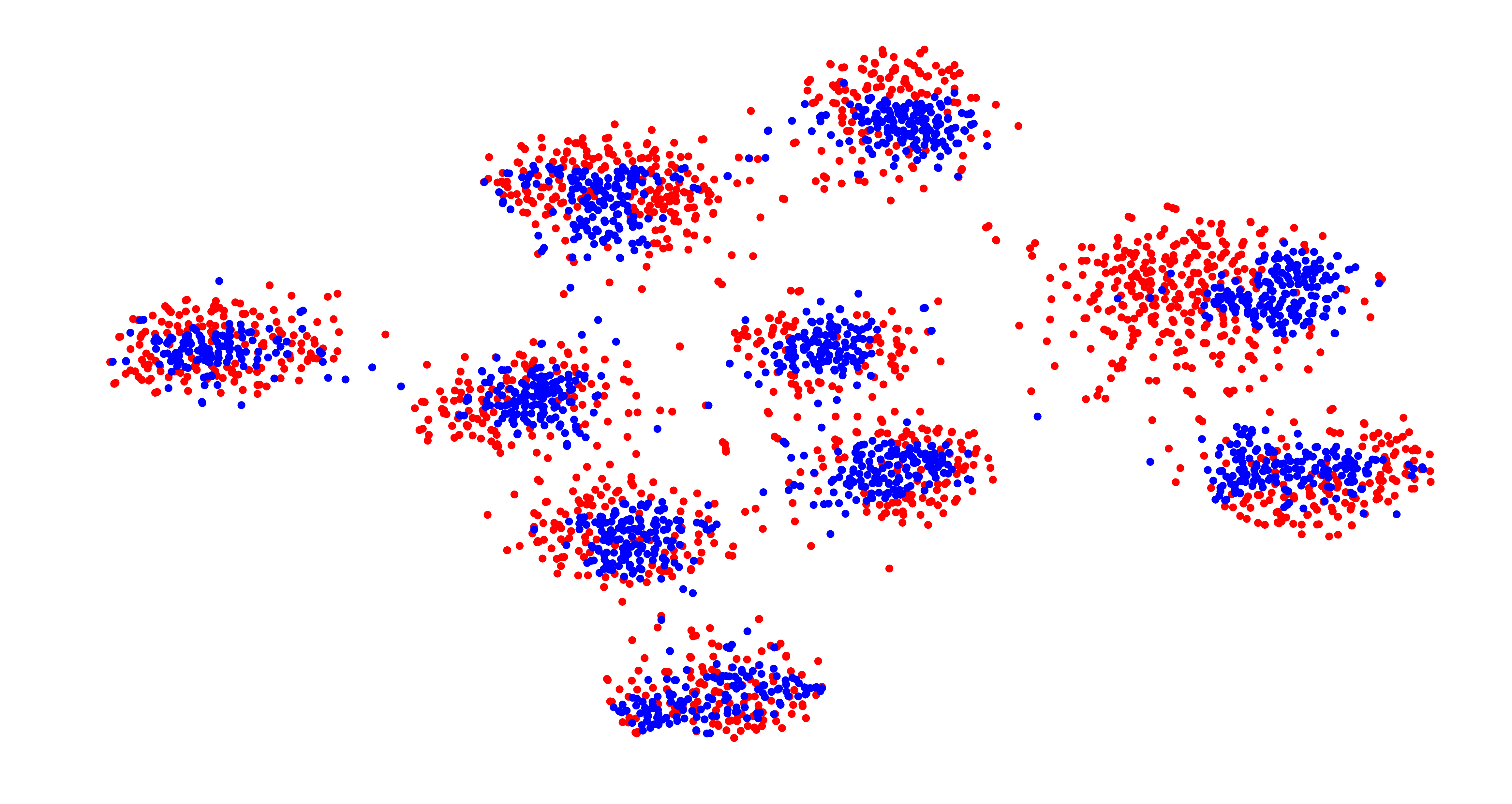}
	 \caption{Task-specific discriminator
	 \label{vis_ours}}
	 \end{subfigure}

	\\
    \begin{subfigure}[b]{0.35\linewidth}
    \includegraphics[clip,width = .7\linewidth,height = .55\linewidth]{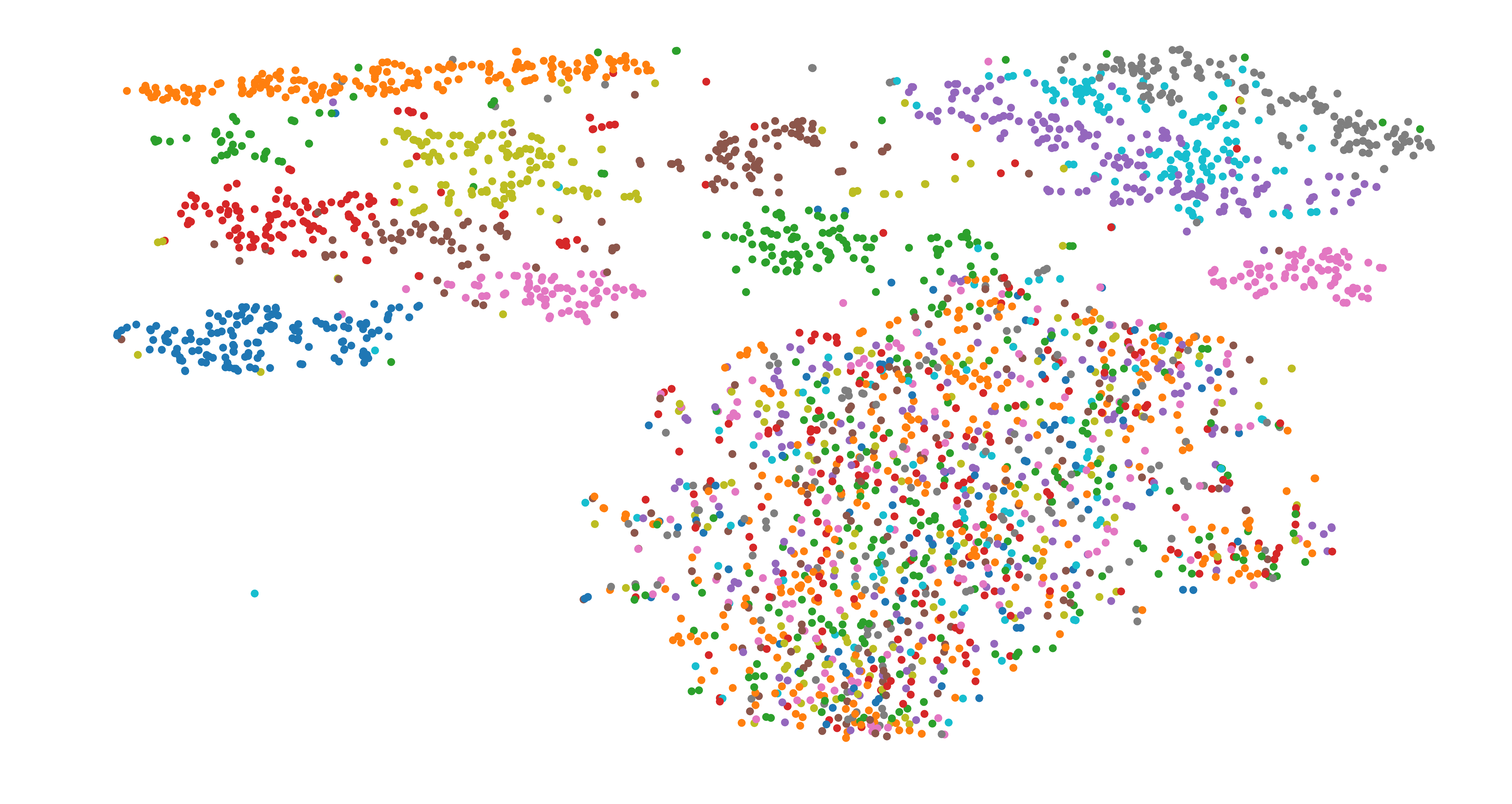}
    \caption{Original (by classes)\label{vis_class}}
    \end{subfigure}
    &
    \begin{subfigure}[b]{0.35\linewidth}
	 \includegraphics[clip,width = .7\linewidth,height = .55\linewidth]{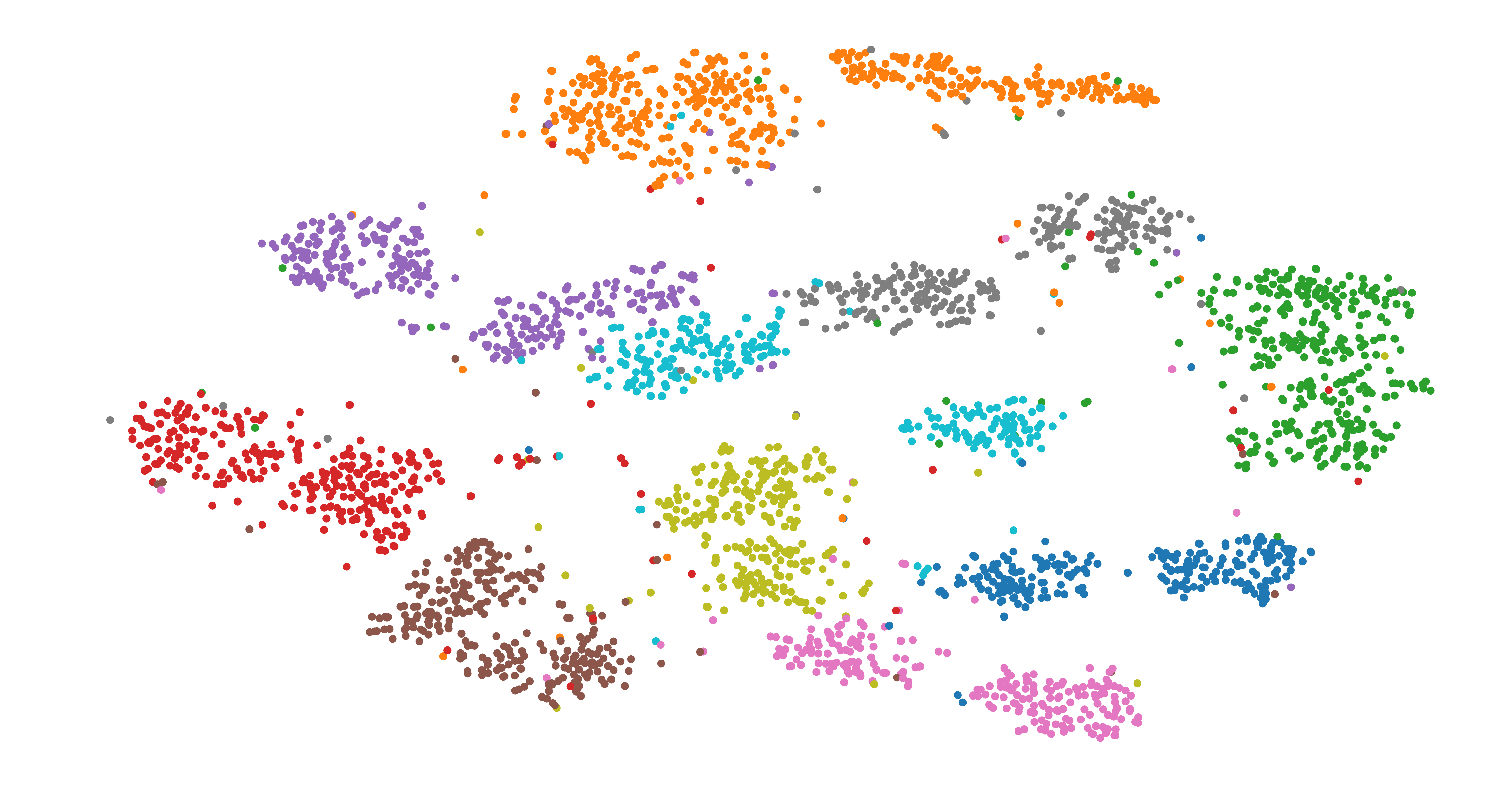}
	 \caption{Adversarial discriminator
	 \label{vis_binary_class}}
	 \end{subfigure}
	 &
	 \begin{subfigure}[b]{0.35\linewidth}
	 \includegraphics[clip,width = .7\linewidth,height = .55\linewidth]{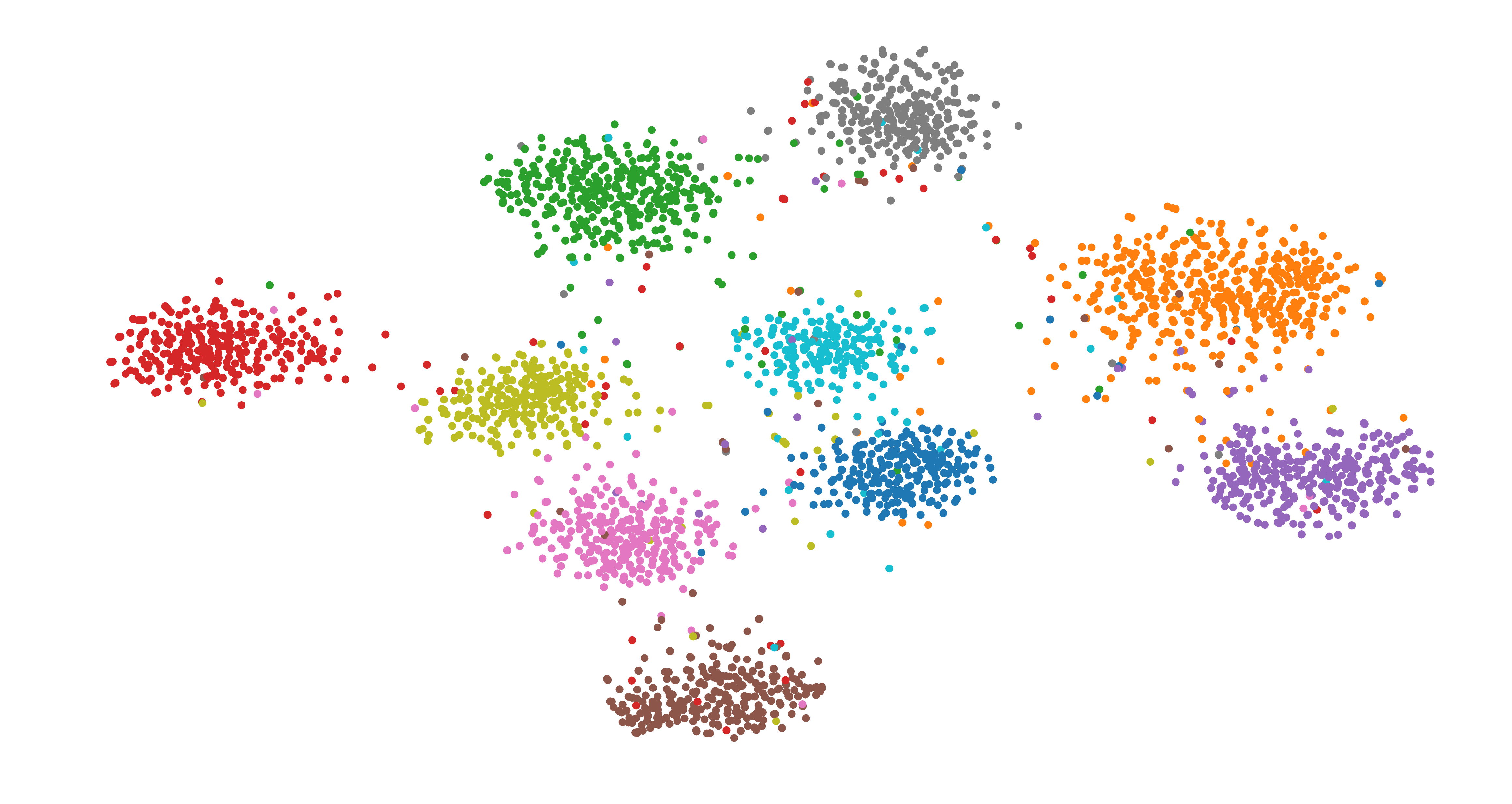}
	 \caption{Task-specific discriminator
	 \label{vis_ours_class}}
	 \end{subfigure}
    \end{tabular}
    \caption{Feature visualization for embedding of digit datasets for adapting \textbf{SVHN} to \textbf{MNIST} using t-SNE algorithm. The first and the second rows show the domains and classes, respectively, with color indicating domain and class membership. \protect\subref{vis_original},\protect\subref{vis_class} Original features.
     \protect\subref{vis_binary},\protect\subref{vis_binary_class} learned features for Ours with (binary) adversarial discriminator. \protect\subref{vis_ours},\protect\subref{vis_ours_class} learned features for Ours with task-specific discriminator.
     }
     \label{visualization}
\end{figure*}
\begin{figure}[b]
\vspace{-1.0em}
\begin{center}
\includegraphics[trim = 0mm 5mm 2mm 0mm, scale=0.25, clip]{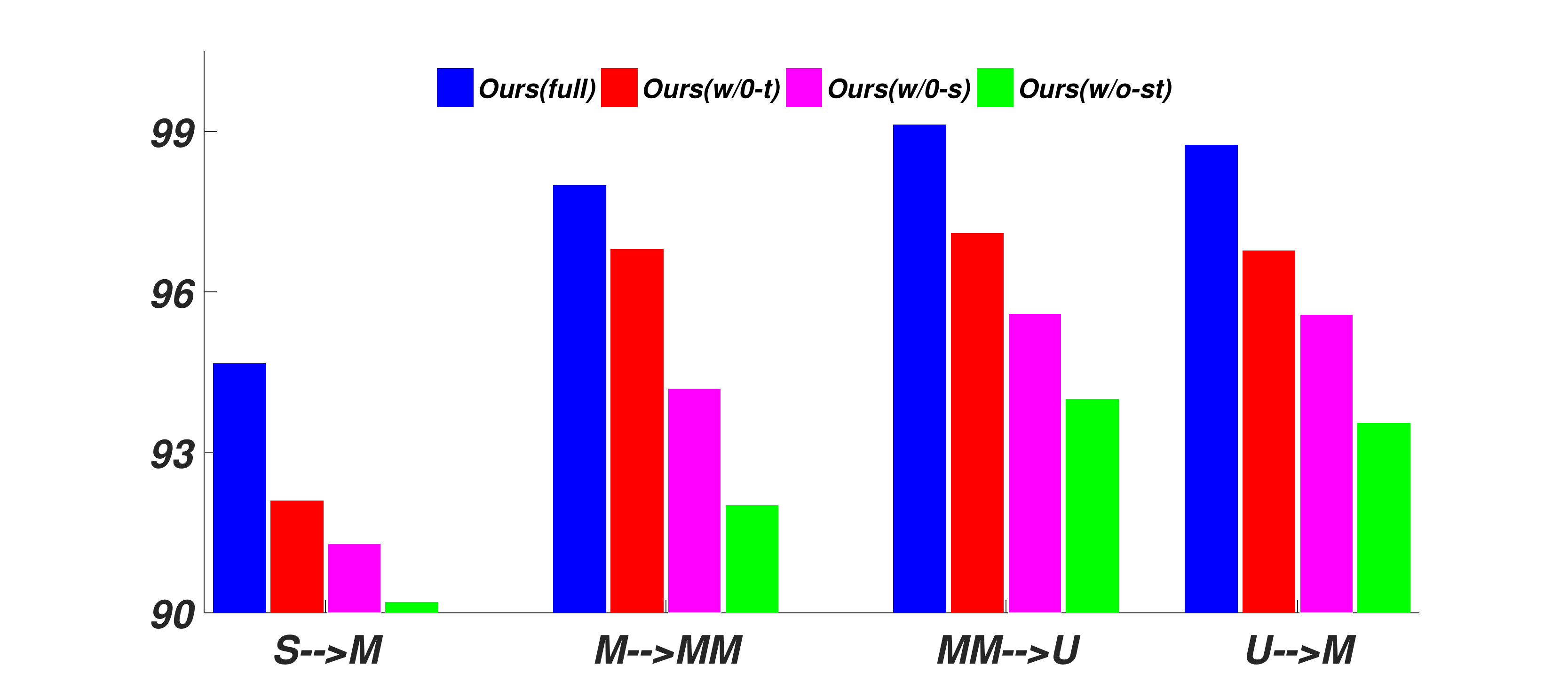}
\end{center}
\vspace{-1em}
\caption{ Ablation of the proposed method on Digit dataset.
The regularization terms contribute to the overall performance.}
\label{ablation}
\end{figure}
\subsection{Analysis of the task-specific discriminator}\label{ex:task}
To measure how effective the new task-specific discriminator is, we conducted an experiment to compare the task-specific discriminator with the standard adversarial discriminator (training a logistic
function on the discriminator by assigning labels 0 and 1 to the source and target domains respectively
and training the encoder with inverted labels). The results are shown in \autoref{ablation2}. As is evident from the figure, there is a substantial increase in accuracy over all adaptation scenarios on switching from the standard adversarial discriminator to our task-specific discriminator. The superiority of the performance is mainly due to explicitly accounting for task knowledge
in the proposed discriminator during adversarial training that encourages the discriminativity of the source/target samples in the feature space.

We further visualize the distribution of the learnt shared features to investigate the effect of task-specific discriminator (\textbf{Task-d}) and its comparison to adversarial discriminator (\textbf{Adv-d}). We use t-SNE~\cite{maaten2008visualizing} on \textbf{SVHN} to \textbf{MNIST} adaptation to visualize shared feature representations from two domains. \autoref{visualization} shows shared features from source (\textbf{SVHN}) and target (\textbf{MNIST}) before adaptation (\subref{vis_original},\subref{vis_class}), after adaptation with \textbf{Adv-d} (\subref{vis_binary},\subref{vis_binary_class}), and after adaptation with \textbf{Task-d} (\subref{vis_ours}, \subref{vis_ours_class}).

While a significant distribution gap is present between non-adapted features across domains (\subref{vis_original}), the domain discrepancy is significantly reduced in the feature space for both \textbf{Adv-d} (\subref{vis_binary}) and \textbf{Task-d} (\subref{vis_ours}). On the other hand, adaptation with \textbf{Task-d} led to pure and well-separated clusters in feature space compared to the adaptation with \textbf{Adv-d}, and leads to superior class separability. As supported by the quantitative results in \autoref{ablation2}, this implies that enforcing clustering in addition to domain-invariant embedding was essential for reducing the classification error. This is depicted in \subref{vis_ours_class}, where the points in the shared space are grouped into class-specific subgroups; color indicates the class label. This is in contrast to \autoref{vis_binary_class}, where the features show less class-specificity.

\subsection{Ablation Studies}\label{as}
We performed an ablation study for our unsupervised domain adaptation approach on Digit dataset. Specifically, we considered training without source regularization, denoted as \textbf{Ours (w/o-s)}, training without target regularization, \textbf{Ours (w/0-t)}, and training by excluding both the source and the target regularization, \textbf{Ours (w/o-st)}.

The results are shown in \autoref{ablation}. As can be seen, removing one or more of the objectives results in noticeable performance degradation. The more parts are removed, the worse the performance is.  More precisely, disabling the source regularizer results in an average $\approx 3.5\%$ drop in performance. That demonstrates that the source regularizer can improve the generalization over target samples by encouraging the source features to be domain-invariant, less informative about the identity of either of the domains.  Immobilizing the target regularizer leads to $\approx 2.0\%$ average drop in performance. These results strongly indicate that it is beneficial to make use of the information from unlabeled target data the during classifier learning process, which further strengthens the feature discriminability in the target domain.  Finally, the average performance drop that stems from disabling both the source and the target regularizer is $\approx 5.5\%$.  This suggests that the two components operate in harmony with each other, forming an effective solution for domain adaptation.

\section{Conclusion}

We proposed a method to boosts the unsupervised domain adaptation by explicitly accounting for task knowledge in the cross-domain alignment discriminator, while simultaneously exploiting the agglomerate structure of the unlabeled target data using important regularization constraints. Our experiments demonstrate the proposed model achieves state-of-the-art performance across several domain adaptation benchmarks.

{\small
\bibliographystyle{ieee}
\bibliography{egbib}
}

\end{document}